\documentclass{article} 

\usepackage{arxiv}

\usepackage[utf8]{inputenc} 
\usepackage[T1]{fontenc}    
\usepackage{hyperref}       
\usepackage{url}            
\usepackage{booktabs}       
\usepackage{amsfonts}       
\usepackage{nicefrac}       
\usepackage{microtype}      
\usepackage{lipsum}		
\usepackage{graphicx}
\usepackage{natbib}
\usepackage{doi}
\usepackage{placeins}
\usepackage{ bbold }
\usepackage{multirow}
\usepackage{subcaption}
\usepackage{amsmath}
\usepackage{amssymb}
\usepackage{booktabs}
\usepackage[table]{xcolor}

\usepackage{bm}
\usepackage{dsfont}
\usepackage{pifont}
\usepackage{ stmaryrd }
\usepackage{tabularx}
\DeclareMathOperator*{\iouk}{IoU_{k}}
\DeclareMathOperator*{\iopk}{IoP_{k}}

\author{
    Karim Radouane$^1$, Andon Tchechmedjiev$^1$, Julien Lagarde$^2$, Sylvie Ranwez$^1$\\ 
    1. EuroMov Digital Health in Motion, Université de Montpellier, IMT Mines Ales, Ales, France\\
    2. EuroMov Digital Health in Motion, Université de Montpellier, IMT Mines Ales, Montpellier, France 
}

\date{}

\begin{document}

\title{Motion2Language, Unsupervised learning of synchronized semantic motion segmentation}

\maketitle

\begin{abstract}
    In this paper, we investigate building a sequence to sequence architecture for motion to language translation and synchronization. The aim is to translate motion capture inputs into English natural-language descriptions, such that the descriptions are generated synchronously with the actions performed, enabling semantic segmentation as a byproduct, but without requiring synchronized training data. We propose a new recurrent formulation of local attention that is suited for synchronous/live text generation, as well as an improved motion encoder architecture better suited to smaller data and for synchronous generation. We evaluate both contributions in individual experiments, using the standard BLEU4 metric, as well as a simple semantic equivalence measure, on the KIT motion language dataset. In a follow-up experiment, we assess the quality of the synchronization of generated text in our proposed approaches through multiple evaluation metrics. We find that both contributions to the attention mechanism and the encoder architecture additively improve the quality of generated text (BLEU and semantic equivalence), but also of synchronization. Our code is available at \url{https://github.com/rd20karim/M2T-Segmentation/tree/main}.
\end{abstract}


\maketitle





\section{Introduction}
Motion-Language processing constitutes an emerging field within computer vision, with evident links to  cognitive science, computational linguistics and knowledge engineering, robotics, social computing. The current prism of study of the interplay between motion and language is mainly focused on preliminary exploratory investigations pertaining to the general association between close descriptions of the movements and its parameters (mainly motion capture). The task of Motion to Language mapping or translation consists of mapping sensor-motion parameters (e.g. MoCap) to natural language descriptions. Currently, few datasets include direct motion-to-description correspondences, enabling the use of supervised machine learning, mainly the KIT Motion-Language Dataset (KIT-ML) \citep{Mandery2016} and HumanML3D \cite{Guo_2022_CVPR}. In this paper, we take interest in motion-to-language mapping, exploring different designs of seq2seq architectures. To our knowledge, ours is the first system for unsupervised learning of motion-to-language segmentation and synchronization. We show why classical seq2seq architectures are unable to learn synchronous correspondences and propose iterative evolutions of the architecture for unsupervised learning of semantic motion segmentation based attention weights. We evaluate the models using the BLEU@4 score, which we extend with a semantic coherence metric. Finally, to measure the performance of segmentation, we propose different methods to compute the evaluation metrics for the synchronicity of semantic segmentation. We manually annotate the timings of motion segments of a subset of the KIT-ML dataset (a portion of our test set).

\subsubsection*{Highlight of main contribution}
\textit{\begin{enumerate}
    \item Quantitative and visual illustration of the limitations of recurrent encoder on alignment.
    \item Introduction of recurrent local 
 attention for synchronous generation of text along the motion.
    \item Inference of motion segmentation using only attention weights.
    \item Proposition of a method to evaluate segmentation results and synchronization performance.
    \item Manual annotation of the timings of motion segments on a subset of the test set, and an additional examination of data set descriptions after automated orthographic and grammatical corrections.
\end{enumerate}}

\section{Related work}

In this section we will cover three main areas of relevance to our direct contributions, namely motion-language datasets and systems (main focus of the paper) and human motion segmentation (the main use-case for our approach).

\subsection{Motion-Language Datasets and Systems}
There are two public datasets mapping human pose sequences to linguistic descriptions: the KIT Motion Language dataset (KIT-ML) \cite{Plappert2016} and the HumanML3D (HML3D) \cite{Guo_2022_CVPR}. The former is build by crowdsourcing annotations on the KIT Whole-Body Human Motion Database, and the latter by expert annotation of a subset of the AMASS dataset. Guo et al. \cite{Guo_2022_CVPR} also introduce an augmentation of KIT-ML using the same process as for HumanML3D.

\begin{table}[ht]
\centering
\begin{tabular}{lcccc}
\toprule
\bf Subset    & \bf Number  & \bf Train & \bf Test & \bf Val. \\
\midrule
\multirow{2}{*}{KIT-ML}& \ motions  & 2375  & 291 & 262      \\ 
                            &   samples & 5071  & 612  & 568 \\
\midrule
\multirow{2}{*}{KIT-ML-aug}& \ motions  & 4886  & 830 & 300      \\ 
                            &   samples & 10408  & 1660  & 636 \\
\midrule
\multirow{2}{*}{HML3D-aug}    & motions & 22068  &  4160  & 1386    \\ 
                              & samples & 66734 & 12558 & 4186 \\
\bottomrule
\end{tabular}
\caption{Training splits dans dataset statistics, for KIT-ML and Human ML3D \cite{Guo_2022_CVPR}. (-aug: after augmentation) }
\label{tab:kit_h3d_size}
\end{table}

The first deep learning approach on the KIT-ML dataset was introduced by Plappert et al. \cite{Plappert2017}. They created a system capable of generating movement from textual descriptions and generating text that describes movement using a bidirectional LSTM encoder-decoder architecture.  Later systems mainly focused on language to motion generation \cite{Lin2018, Ghosh2021, petrovich22temos}, but motion to language generation has seen a resurgence with the introduction of HumanML3D. 

Gotsu et al. \cite{Goutsu2021} explore adversarial seq2seq architectures for motion-to-language generation, achieving SOTA on KIT-ML. Guo et al. \cite{chuan2022tm2t} take a bidirectional motion quantization based transformer approach, and are the first to use HumanML3D for motion-to-language generation. Although they achieve SOTA for motion generation on HumanML3D (the first system evaluated on the dataset). The transformer based method have not led to higher result on reverse direction (motion2text) specifically on KIT-ML (BLEU4 $18.4\%$).

\subsection{Human motion segmentation}

 There is no unified and consensual definition of motion segmentation, here we will focus specifically on approaches on skeleton data, which is what interests us. Lin and Kuli\`c \cite{Lin-seg-2014} identify and segment movement repetitions using velocity features and stochastic modeling to select motion segment candidates, then, they apply a Hidden Markov Model (HMM) to select the final segment. Kili\`c et al. In the same vein, Dana et al. \cite{dana-2009} learn to extract human motion primitives (\textit{Lower arm raise}, \textit{Bow Up/Down},...) based on HMMs from a continuous time-series data converted to joint angles. Mei et al. \cite{Mei2021} consider segmentation from the angle of the unsupervised segmentation of human limb motion sequences using an autoregressive moving-average model (ARMA) to fit each limb bone angle, and defining segment points as the position where ARMA predictions no longer satisfy goodness of fit. A technique based on a collaborative representation of 3D skeleton sequences was proposed by Li et al. \cite{Li2018}, where Aligned Cluster Analysis (ACA) \cite{ACA_zhou_2008} and its Hierarchical (HACA) variant \cite{HACA_zhou_2012} are used to identify sequences. 
 More recent works \cite{Ma2021, Filtjens2022} rely on spatial temporal graph convolutions networks. But to our knowledge, no previous work was done in the aim to perform motion segmentation by synchronizing text generation to human skeleton motion.

\section{Methods}

\begin{figure*}[h]
\centering
\includegraphics[width=0.8\textwidth]{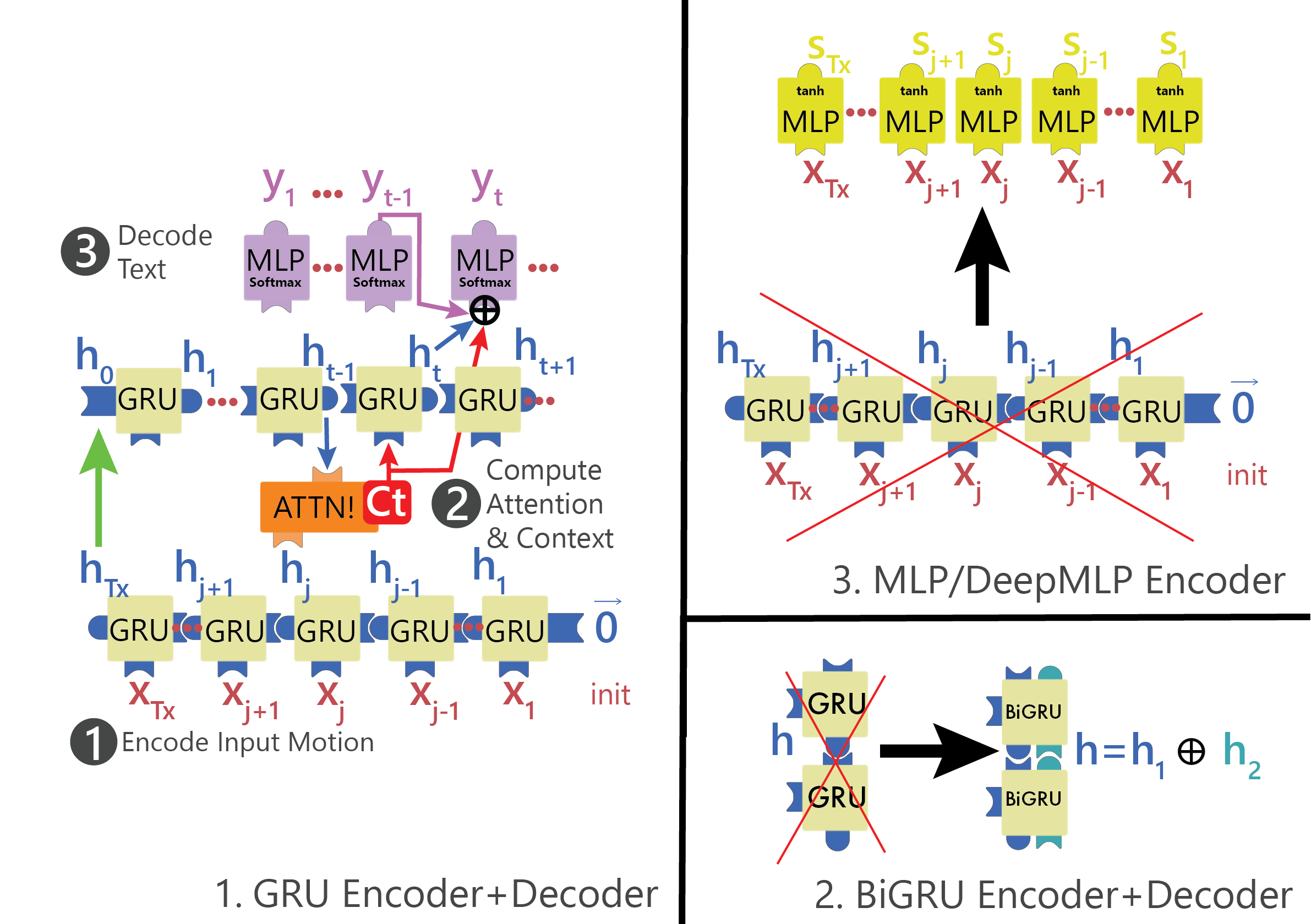}
\caption{Considered Architectures.}
\label{fig:all_arch}
\end{figure*}

Motion to language generation consist in generating a natural language description \(y\) of a motion sequence \(x\), which can be seen as a sequence to sequence (seq2seq) mapping task, and more specifically a multi-modal Neural Machine Translation (NMT) system. In this paper, we propose a model that goes beyond simple mapping, by adapting NMT architectures to generate text aligned with motion. We explore both:

\begin{itemize}
    \item \textbf{Architectural variations}  The first variant use the GRU as motion encoder, the second variant use the Bidirectional GRU. For the third variant we propose to use a frame-level feature extractor based on the MLP that forgoes source summarization. Fig.\ref{fig:all_arch} gives an overview of the considered architecture variations. 
    \item \textbf{Attention mechanism} We consider soft and local attention baselines, and our proposed recurrent local attention tailored for unsupervised synchronous text generation.
\end{itemize}

\begin{figure*}[h]
\centering
\includegraphics[width=0.9\textwidth]{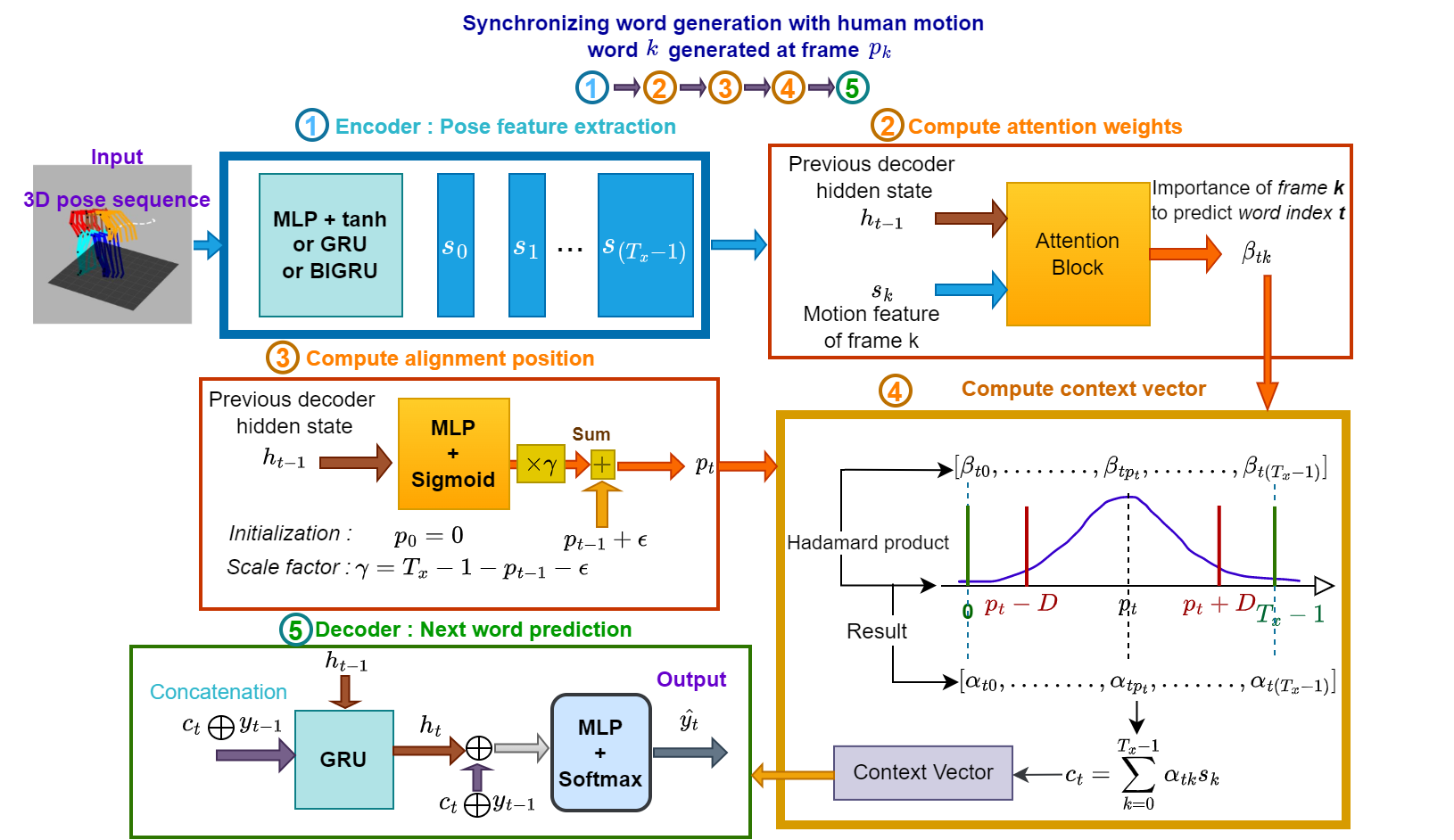}
\caption{Phases of motion to language generation.}
\label{fig:all_arch_seg}
\end{figure*}

As illustrate in the Fig. \ref{fig:all_arch} We experiment three different motion encoders GRU, Bi-GRU and MLP with a GRU decoder. The \(x_j\) vector is the raw pose data of human skeleton at frame \(j\). The output of the encoder frame \(j\) and the output of the decoder GRU cell \(h_{t-1}\) are used to compute attention weights and in turn the context vector \(C_t\) at token \(t\). The previous word embedding \(y_{t-1}\), the hidden state of the GRU \(h_t\) and the context vector \(C_t\) are concatenated and given as input to the final MLP/softmax layer. This layer predicts a probability distribution over the vocabulary of the language model. At each time step, the word with maximum probability, \(y_t\) is generated using a greedy decoding approach. This token is fed into the GRU for the next time step, and the process is repeated until the end token \(<\mbox{eos}>\) is reached. The Fig.\ref{fig:all_arch_seg} specify the phases order of language prediction with respect to attention weights $\alpha_{it}$ used for segmentation.

In the remainder of the section, we discuss the classical NMT architectures and the role of soft and local attention in the computation of the sequence alignment (Section \ref{sec:baselines_classical}), and outline the main limitations of soft and local attention if we want to achieve synchronous generation (Section \ref{sec:baselines_limitation}). We then introduce local recurrent attention, a novel formulation that allows synchronicity during generation (Section \ref{sec:introd_rec_att}). Finally, we present an evolution of the encoder architecture better suited for small data and improving on synchronous generation (Section \ref{sec:rethink_seq2seq_arch}). 

\subsection{Baseline NMT Architecture and limitations}
\label{sec:baselines}
We first briefly present the baselines and their formulation, followed by a detailed analysis and discussion of our proposed evolution of the architecture to enable better synchronous generation. In this work, for all baseline variants, we use a canonical NMT architecture \cite{Sutskever2014} adapted to receive poses as input instead of the source language tokens. The model is a Seq2seq Encoder/Decoder network, where the Encoder and Decoder are based on a Gated Recurrent Unit (GRU). The architecture variants are trained to model the probability \(p(y\mid x) = \prod_{t=1}^{\mid y \mid } p(y_t \mid y_{<t},x)\) through the minimization of the cross-entropy loss.

\subsubsection{Classical global formulations of attention}
\label{sec:baselines_classical}

\paragraph{Soft attention formulation}
The first type of attention considered is soft attention, the cannonical formulation for NMT presented by Bahdanau et al. \cite{Bahdanau2014}. The attention coefficients \(\alpha_{ij}\) (Equation \ref{eq:soft_att_coeff}) are computed over the energy of possible alignments (Equation \ref{eq:soft_att_energy}). 
\begin{align}
    \alpha_{ij}&=softmax([e_{i1},\cdots,e_{iT_x}]) \label{eq:soft_att_coeff} \\
    e_{ij} &= v_a^{\top} \tanh\left( W_a h_{t-1} + U_a s_j \right) \label{eq:soft_att_energy}
\end{align}

Here, \(T_x\) is the length of the motion sequence, $h_{t-1} \in \mathbb{R}^{n'} $ the decoder hidden state at word \(t-1\), \(s_j\) the encoder output for motion frame \(j\), while $v_a \in \mathbb{R}^{n'\times 1}, W_a \in \mathbb{R}^{n'\times n}$ and $U_a \in \mathbb{R}^{n'\times  m}$ are learnable parameters, with $m=1$ for the unidirectional GRU and MLP-based encoder, and \(m=2\) for the bidirectional GRU.

\paragraph{Local attention formulation}
The soft attention also allows the decoder to have access to information about past and future frames, and doesn't incentive the model to learn a precise localization the motions that corresponds to a particular set of words. The model learns to predict the moment when maximal information is available (at the very end). One solution can be to limit the information visible to the decoder through local attention \cite{Luong2015}, applying a window of weights that only retains a local interval around the current frame. 
Luang et al. \cite{Luong2015} propose to learn an aligned position $p_t$ for each decoding step $t$. A window centered around the source position $p_t$ is used to compute the context vector $c_t$. The sum is directly computed inside the window $\llbracket p_t-D, p_t+D \rrbracket$~\footnote{$\llbracket i, j\rrbracket$ denotes set of integers between $i$ and $j$ both included. We will use the Bourbaki convention throughout the paper.}, where $D$ is a hyperparameter. For the computation of \(p_t\), we use $h_{t-1}$ instead of $h_t$ as formulated by Bahdanau et al. \cite{Bahdanau2014} as show in Equation \ref{eq:align_pt_local}. 

\begin{align}\label{eq:align_pt_local}
    p_t = T_x.\sigma({{v}_p}^{T}\tanh(W_{p}h_{t-1}))
\end{align}

Where, $\sigma(.)$ denotes the sigmoid function, and where $W_{p} \in \mathbb{R}^{n' \times n'} $ and ${v}_p \in \mathbb{R}^{n'\times 1}$ are learnable parameters. In this work we consider a gaussian windowing function as it leads to smoother generation (see Equation \ref{e:align_p}).

\begin{equation}
        a_{ij} = \alpha_{ij} \cdot \exp({-\frac{(j-p_t)^2}{2r^2}}) 
        \label{e:align_p}
\end{equation}

In all the following preliminary experiments we note by \textit{Soft.att}, \textit{Local.att}, \textit{Local.rec.att} respectively the soft attention, local attention and local recurrent attention. The BLEU refers to the default BLEU@4 score. In all attention map Figures, the red rectangles represent the location of maximum attention for each word along the frame axis. The orange rectangles delimit the range of above zero activation.

\subsubsection{Limitations of local and soft attention}
\label{sec:baselines_limitation}

 In order to observe the behavior and limitations of these classical formulations of attention and draw meaningful avenues of improvement, we perform a preliminary experiment, whereby we train the baseline GRU architecture (GRUenc + GRUdec), both unidirectional and bidirectional, on cartesian coordinate features, following the procedure described in Section \ref{sec:train_strat}, and plot the attention maps on a few representative motion samples.

\begin{figure}[ht]
    \centering
    \begin{subfigure}[b]{0.45\linewidth}
        \centering
        \includegraphics[width=\columnwidth]{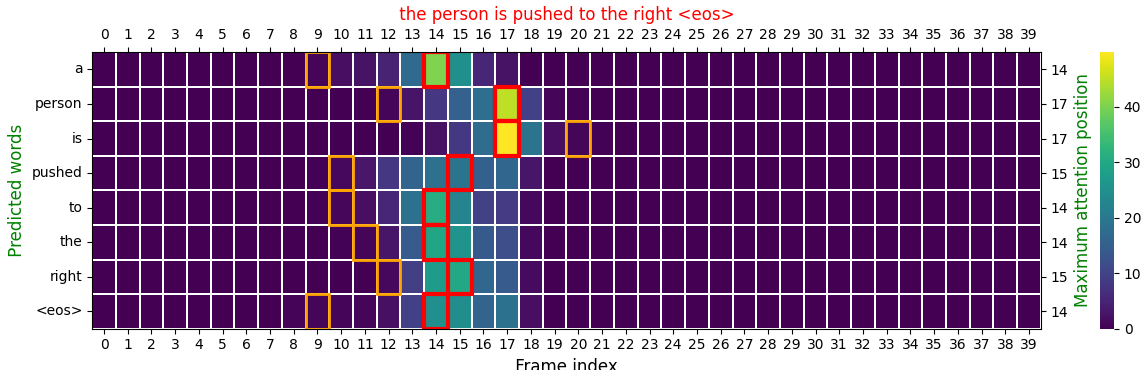}
        \caption{BiGRU-cartesian[Soft.att] : Pushing action in the range $\llbracket11,20\rrbracket$.}
        \label{fig:bigru_att_map}
    \end{subfigure}
    \hfill
    \begin{subfigure}[b]{0.45\linewidth}
        \centering
        \includegraphics[width=\columnwidth]{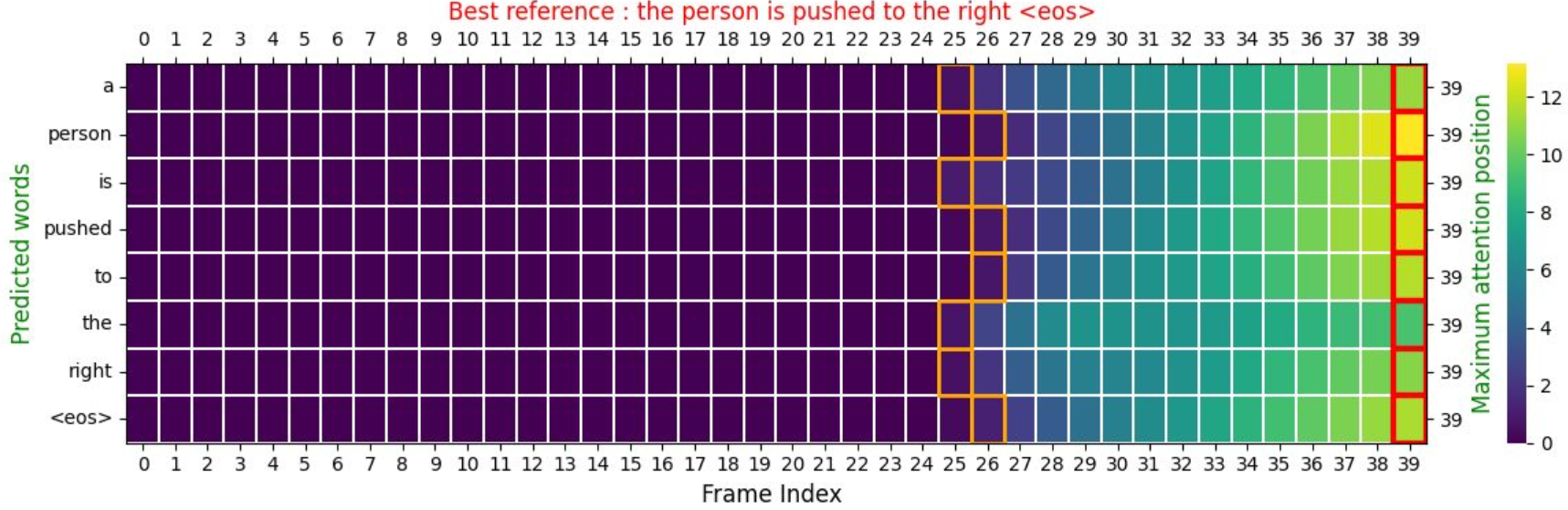}
        \caption{GRU-cartesian[Soft.att] : Pushing action in the range $\llbracket11,20\rrbracket$.}
        \label{fig:uni_cart_soft_att}
    \end{subfigure}
    \vfill
    \begin{subfigure}[b]{0.45\linewidth}
        \centering
        \includegraphics[width=\columnwidth]{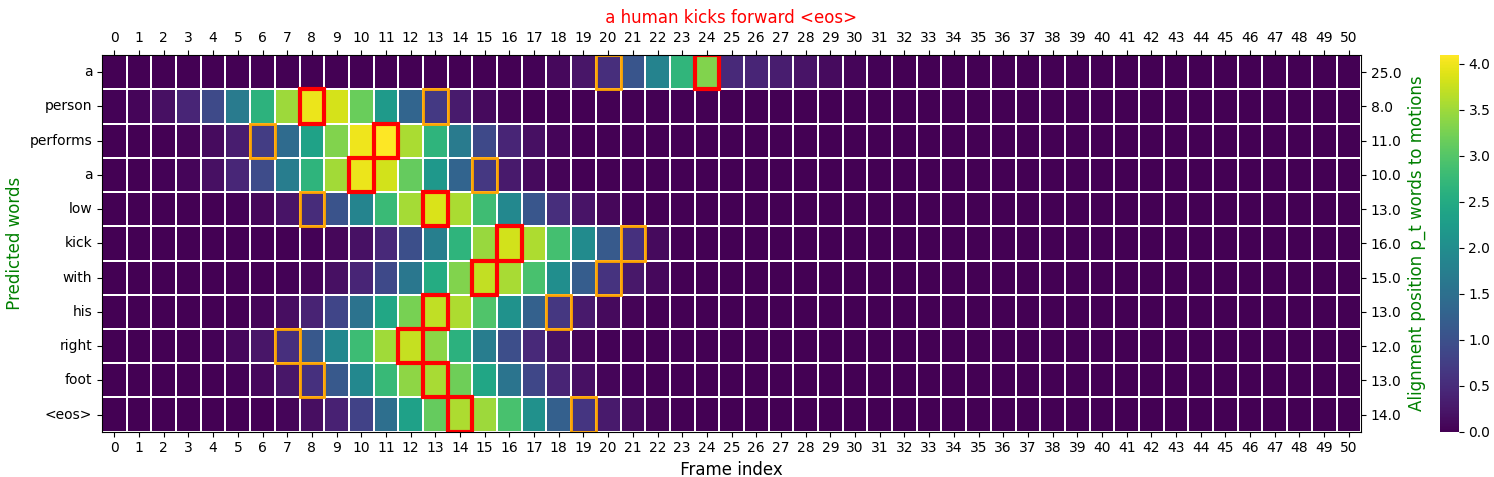}
        \caption{BiGRU-cartesian[Local.att Mask False] : Kicking action in the range $\llbracket12,24\rrbracket$.}
        \label{fig:Bi_cart_local_att_maskFalse}
    \end{subfigure}
    \hfill
    \begin{subfigure}[b]{0.45\linewidth}
        \centering
        \includegraphics[width=\columnwidth]{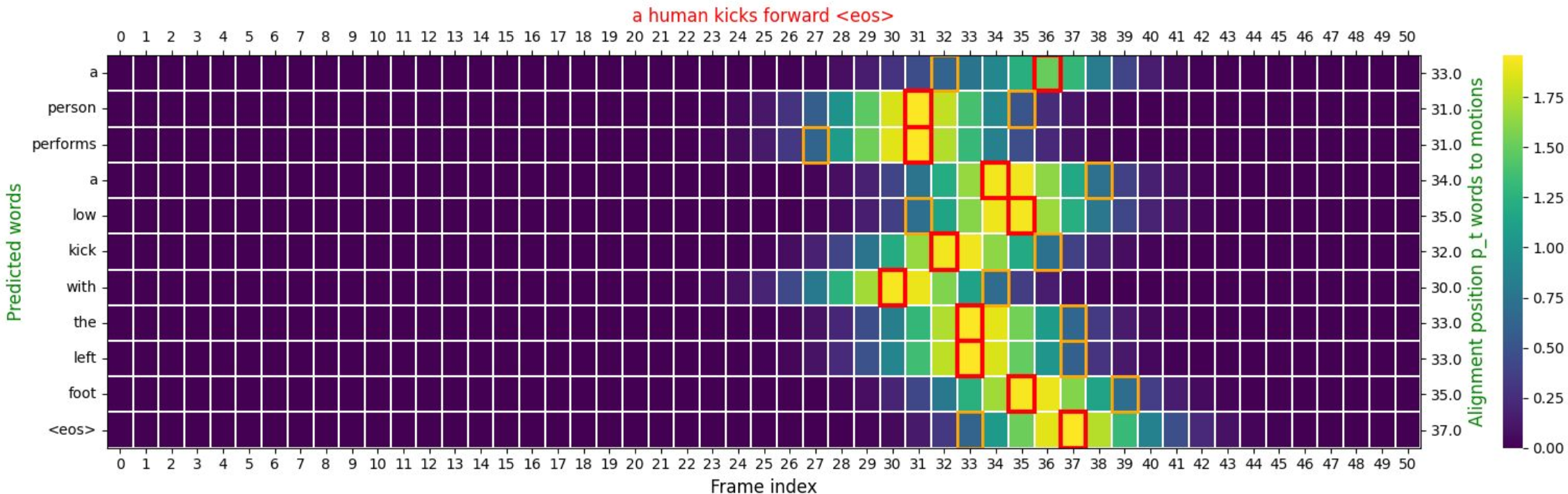}
        \caption{GRU-cartesian[Loc.att] : Kicking action in the range $\llbracket12,24\rrbracket$.}
        \label{fig:unigru_local_maskFalse_kick_28}
    \end{subfigure}
\end{figure}

\label{sec:limit_loc_soft}

\textit{Soft attention} For BiGRU-Cartesian as we see attention in the Fig.\ref{fig:bigru_att_map} we have a disorder in the position of maximum attention, which doesn't allow a synchronized generation of words. At this stage, the attention gives only correct information about the range of motion $\llbracket12,21\rrbracket$  and action description.
For the same sample in the case of unidirectional GRU, the maximum  attention (Figure \ref{fig:uni_cart_soft_att}) was localized at the end frames. Observing the attention maps of the test set, observe that in general the maximal attention distribution was localized around the frame marking the end of the movement when the action is completed.

\textit{Local attention}
For BiGRU-cartesian, applying the mask results in the alignment positions being equal to the length of the motion. On the contrary, when the mask wasn't applied, as illustrated by the Fig.\ref{fig:Bi_cart_local_att_maskFalse}, surprisingly the alignment positions were well distributed in the range $\llbracket5,24\rrbracket$, which corresponded to the interval of motion execution, but fell short of being well synchronized: for the word describing the action \textit{“kick”} attention weights were correctly distributed in the range $\llbracket13,19\rrbracket$ around the maximum attention frame $16$ (center time of kick motion). For the same sample, in the case of the unidirectional GRU, the attention weights have a similar distribution, but in the range $\llbracket25,41\rrbracket$, after the end of the action. In most cases, for the unidirectional GRU, the attention positions were distributed in the correct zones along the time/frame axis but not well synchronized.

\textbf{Constraint on alignment position for synchronization}
\label{par:constraint_align}

In order to obtain synchronous generation using only, $\bm{p_t}$ we derive the idea of calculating this position under the constraint that $\bm{p_{t-1}\leq p_t}$. Although this constraint is language dependent and not universally true at the word level, it does hold up at the phrase level. For example, the words $\{"the", "a", "person"\}$ are not related to the monotony of frame generation, but for action words like (“walk”, “jump”), the succession  'walk'  then 'jump' happens successively in time, as results the word describing these actions will also appear successively in the human description. The description is made step by step as the motion is generated. The positions $p_t$ will be more interesting for words that describe actions. The other intermediate positions, for language connection, will serve as a relative position from which the next action is localized. When we have simultaneity of two actions described respectively by word $w_i$ and $w_j$, ideally,  the model will learn to predict approximately same position alignment $\bm{p_{i}\approx p_{j}}$. Meaning that the position $p_t$ will remain approximately constant in the range $\llbracket i, j\rrbracket$. On the side of language generation, the words connecting the two action words will be independent of motion. \textit{The main goal is to associate every set of words in the sentence describing one action to the relevant set of frames based on $p_t$ and attention weights.}

\subsection{Introducing local recurrent attention}

\label{sec:introd_rec_att}

In local attention, the position $p_t$ depends only on the previous hidden state $h_{t-1}$, which is problematic in our case, we observe that the network simply learns to almost always compute $p_t$ at the start, middle, or at the end of the motion.
The BiGRU is biased more towards the start and middle of the motion in Fig.\ref{fig:Bi_cart_local_att_maskFalse} (human movement end at the frame $27$), while a unidirectional GRU favoured the positions after the end of the human movement in Fig.\ref{fig:unigru_local_maskFalse_kick_28} (same sample). Intuitively, the network gleans on compressed information about the entirely of the sequence and issues a prediction when the amount of information is maximal on average, thus negatively impacting its ability to perform synchronous predictions. Indeed, the network chooses the path of least effort without searching along the source length for finding the most relevant motion interval to output for a given phrase. 

\subsubsection{Formal definition}
\label{sec:introd_rec_att_formal_def}
To overcome this limitation, we want the model to associate a subset of successive frames to the most relevant word(s) for an exact alignment and synchronization of segments of human motion with the generated text, and to this effect we propose a new formulation to compute the position $p_t$ in Equation \eqref{eq:pt_rec}\footnote{"$\cdot$” is the scalar product}, enforcing the important constraint that $\bm{p_{t-1}\leq p_t}$ as explained in paragraph \ref{par:constraint_align}. 

\begin{equation} \label{eq:pt_rec}        
     p_{t} = p_{t-1} +\epsilon + (T_x-1-p_{t-1}). \sigma(v_p^t\cdot W_{p}h_{t-1})
\end{equation}

The scaling factor $\bm{(T_x - 1 - p_{t-1})}$ maintains the position $p_t$ within the range of the motion frames $\llbracket 0,T_x-1 \rrbracket$. The position $p_t$ indicate where to look in the source frames at a time $t$. The  shift position $\bm{p_{t} - p_{t-1}}$ depends on the previous hidden state of the decoder $h_{t-1}$. 

The learned motion segment $S_{t}$ is delimited from the global sequence through masking \footnote{equivalent to applying a truncated Gaussian kernel} as defined by Equation \eqref{eq:seg_def}. 
Then, using the recurrence relation in Equation \eqref{eq:pt_rec} and the formulation of $S_{t}$ in Equation \eqref{eq:seg_def}, we can derive intersection segment boundaries as concluded in Equation \eqref{eq:seg_intersc}. As mentioned previously, we use the Bourbaki notation for intervals, where $\llbracket i,j\llbracket$ refers to the set of integers between $i$ and $j$ ($i$ included, $j$ excluded).

\begin{align}
S_{t} &= \llbracket p_t-D,p_t+D\llbracket \cap \llbracket 0,T_x\llbracket =   \llbracket \max(0,p_t-D),\min(T_x,p_t+D)\llbracket
 \label{eq:seg_def}
\end{align}

\begin{equation}
 S_{t} \cap S_{t-1}  = \llbracket p_{t-1}-D+\epsilon+(T_x-1-p_{t-1}).\sigma(v_p^t\cdot W_{p}h_{t-1}), \min(T_x,p_{t-1}+D)\llbracket 
 \label{eq:seg_intersc}   
\end{equation}

 We define $\bm{\epsilon} \in \mathds{N}$ as a parameter that can be used to force a minimum shifting between two successive positions and also reduce overlapping between successive local attention windows, by setting $\bm{\epsilon = (1-\alpha).2D}$, where $\alpha \in [0,1]$ is an explicit parameter to control overlapping proportion between successive motion segments $ S_{t} \cap S_{t-1}$. 
 
 Using Equation \eqref{eq:seg_intersc} we have $\alpha=0 \Rightarrow S_{t} \cap S_{t-1} =\emptyset$. This can be convenient to ensure proper word ordering and fast training in motions with clearly separate motion segments. Of course, in more complex cases such as simultaneous events, this mechanism is likely insufficient on its own. In addition, as a result of the recurrence effect, $\bm{p_t}$ is mechanically incremented by $\bm{\epsilon}$, meaning that $\bm{p_t \geq \epsilon \times t}$. This can lead the network to miss critical time intervals and to quickly go towards the final positions, particularly in very short and quick motions. To allow generating words describing simultaneous actions at the same time we set $\epsilon = 0$.
 
 Also note that Equation \eqref{eq:pt_rec} implicitly assumes that the motion and language description are \textit{monotonically} related, meaning that the description of successive phases of a movement, will be in chronological order of the performance of the motion. For parameter $D$, we attempted to make it learnable through various formulations, but the values ended up being too high to allow for a correct segmentation in most cases, even when using a boundary function. However, when selecting its value as a hyperparameter we found a constant value for $D$ leading to consistently better results.

\textbf{Context vector} Noting by $s_x(j)$ the source hidden state at time $j \in \llbracket 0,T_x-1\rrbracket$  for a given motion $x$, the context vector is calculated for the two cases using Equation \eqref{eq:c_t_mlp}.

    \begin{equation}
    c_t = \begin{cases}
    \sum_{j=0}^{T_x-1} a_{tj} s_x(j) &\text{Mask is False}\\
    \sum_{j \in T_{x_t}} a_{tj} s_x(j) &\text{Mask is True}
    \end{cases}
    \label{eq:c_t_mlp} 
    \end{equation}
    
\subsubsection{Applying recurrent attention}

Again, in order to observe the behavior of our new formulation with regard to the classical formulations, we perform a second preliminary experiment, whereby we train the baseline GRU architecture (GRUenc + GRUdec), both unidirectional and bidirectional, following the procedure described in section \ref{sec:train_strat}, and plot the attention maps on a few representative motion samples.

\begin{figure}
    \centering
    \begin{subfigure}{0.48\linewidth}
        \includegraphics[width=\columnwidth]{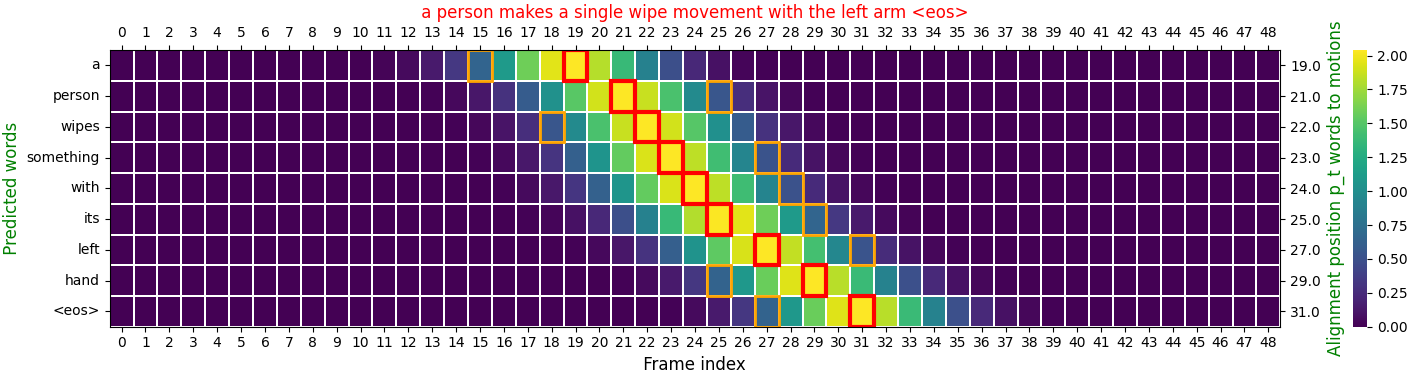}
        \caption{GRU-cartesian[Local.rec.att Mask False]: Wiping motion in the range $\llbracket 11,32 \rrbracket$.}
        \label{fig:Uni_cart_mask_false_loc_rec_1}   
    \end{subfigure} \hfill
     \begin{subfigure} {0.48\linewidth}
        \includegraphics[width=\columnwidth]{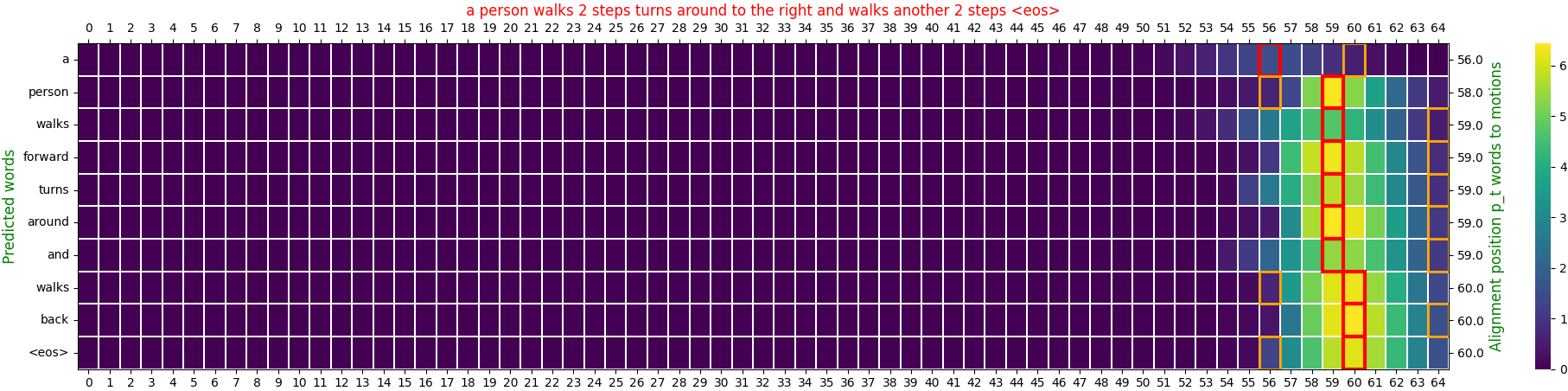}
        \caption{GRU-cartesian[Local.rec.att Mask False]: walk forward in the range $\llbracket 15,38\rrbracket$, turn at frame $40$ , walk backward in $\llbracket 42,53 \rrbracket$.}
        \label{fig:Uni_cart_mask_false_loc_rec_2}
    \end{subfigure}
\end{figure}

Based on the attention maps for both cases Bi-GRU/GRU when applying the mask, the positions of maximum attention are always near the end of the motion sequence. On the contrary, when not applying the mask for the BiGRU, the attention weights were distributed progressively near the end of frames. However, for the unidirectional GRU we begin observing a progressive alignment that start at the end of the motion for some samples like in Fig.\ref{fig:Uni_cart_mask_false_loc_rec_1}, the synchronization was more correct. But in all cases when we have more than one motion or action, like in Fig.\ref{fig:Uni_cart_mask_false_loc_rec_2}, we see no synchronization between motion and language generation. This will be further explained and addressed in section \ref{sec:rethink_seq2seq_arch} as well as the limitations related to this synchronization, as highlighted in section in \ref{sec:limit_loc_soft}.

\subsection{Rethinking feature extraction for synchronous text generation}
\label{sec:rethink_seq2seq_arch}

\textit{Recurrent effect of GRU.} In the inspection, the alignments resulting from the GRU-based motion encoders with various attention formulations in the preliminary experiments, lead to an imperfect synchronization. We hypothesize that this is due to the recurrence effect present in the design of the $GRU$ on the side of the encoder, its hidden state $s_t$ depends on all previous states, and we can write $s_t = \varphi(s_{t-1},x_{t})=\varphi(\cdots \varphi(s_{t-\overrightarrow{k}},x_{t-\overrightarrow{k}}))$. We note by $\varphi$ the classical GRU model Equation and by $\overrightarrow{k},\overleftarrow{k}$ respectively the limit level of compression without important information loss for the forward and backward direction. 
Hence, in the unidirectional case, $s_t$ already encodes all the necessary information about the $\overrightarrow{k}$ previous steps $\{s_{t-\overrightarrow{k}},s_{t-\overrightarrow{k}+1},\cdots, s_{t-1}\}$. Putting maximum attention on $s_t$ is equivalent to \emph{seeing} the $\overrightarrow{k}$ previous motion frames: an action that happens at around frame time $t-\overrightarrow{k}$ can hold relevant global compacted information pertinent for frame $t$. The following clearly explains the misaligned position and late detection of primitive motion, as was illustrated by Figures \ref{fig:uni_cart_soft_att} and \ref{fig:Uni_cart_mask_false_loc_rec_2}. For the Bi-GRU case, the analysis is similar, except that $s_t$ now encodes all the necessary information about the $\overrightarrow{k}$ previous steps and $\overleftarrow{k}$ next information, this explains the localization of attention weight distribution towards the middle of motion segments (See Fig.\ref{fig:bigru_att_map}). Following this rationale, we have determined that a frame-level feature extractor could resolve the late detection problem. Frame-wise feature extraction will push the network to use all source sequence features separately. The adoption of this type of feature extraction, matches similar choices in other works pertaining to alignment tasks, such as sign language/movie subtitle transcription  \cite{Bull2021,Gul2021}. Although, the choice isn't supported by quantitative evidence. The main difference in our approach is that we do not have a supervised timing annotation of movement phases/primitives. Hence, the main innovation in our work (see Equation \eqref{eq:pt_rec}) is the ability of inferring a synchronous alignment only through attention weights using positions $p_t$.

\textit{Motion encoder} To avoid overfitting in the experimented relatively small datasets, we use a low paramter count MLP (Multi-Layer Perceptron) as the frame level feature extractor. Given a motion $x \in \mathbb{R}^{T_x\times63}$, the MLP encodes the sequence motion on a per-frame basis and produce the encoder outputs $s_x \in \mathbb{R}^{T_x \times d_{enc}}$ where $d_{enc}$ is the dimension of the encoded vector. Here, $d_{enc}$ is the same as the decoder hidden size $(d_{enc}=n')$. 

The MLP network consists of $k$ successive non-linear layers as defined by Equation \eqref{eq:mlp_enc}, where $f_i(x(t))=\tanh(W_i.x(t)+b_i)$. The learnable parameters are $W_i \in \mathbb{R}^{63\times d_{i}}$ and $ b_i \in \mathbb{R}^{d_{i}}$, $\tanh$ is the hyperbolic tangent, and where $L$ is the number of layers. 
 
\begin{equation}
\centering
    s_x(t) = (f_0\circ f_1 \circ \cdots \circ f_{L-1})(x(t)) \hspace{8mm} \forall t \in \llbracket 0,T_x-1 \rrbracket
\label{eq:mlp_enc}
\end{equation}

\section{Training strategies}
\label{sec:train_strat}
In previous works by Plappert et al. \cite{Plappert2017} and Inamura et al. \cite{Goutsu2021} the Cartesian coordinates are not used, but we hypothesize that their use is more suitable to generate descriptions pertaining to the global motion and its automatic segmentation.

\textbf{Loss definition}
We define loss as the usual cross entropy by the Equation \eqref{eq:loss}, where $B$ is the set of training examples within in a given batch, for training we use Adam optimizer with an initial learning rate $10^{-3}$. To train, we use Teacher forcing strategy using a ratio of $0.5$ that was empirically selected.

\begin{align}\label{eq:loss}
    Loss &= - \frac{1}{ \mid B \mid}\sum_{(x,y) \in B} \frac{1}{ \mid y \mid^\beta}\sum_{k=0}^{\mid y \mid-1} log(P(y_k\mid y_{<k},x))
\end{align}

\textbf{Word Embedding} We choose to learn the embedding of words in the training, instead of using a pre-trained model. Usually, pre-trained models produce high dimensional vectors, increasing the number of network parameters and leading to over-fitting for small datasets, which isn't adapted in our context.

\textbf{Hyperparameter tuning} In all experiments, we use grid-search to optimize hyperparameters using ray tune \cite{Liaw2018}. We found that in most cases the best choice for better generalization was lower embedding size (64) and hidden size (64) with number of layers (1), we use also the teacher force strategy with a $0.5$ ratio. These parameters are consistent with the small size of the KIT-ML dataset.  We run experiments for $L \in \{2,4\}$ and $d_{i} \in \{64,128,256\}$ except for the final layer $(i=L-1)$ where the output dimension $d_i$ is fixed as $d_i=d_{enc}$.

\section{Experimental Evaluation and Results}
\label{sec:result}
The preliminary experiments and observations give us some insights into the effect of various architectural and formulation choices, but a formal and systematic validation of the proposed systems is indispensable. 
In this section, we first discuss the typical metrics for text generation systems and propose a simple \emph{semantic similarity score} to complement the standard BLEU4 metric. Subsequently, in Experiment 1, we study the effect of the classical soft and local attention formulations compared to our proposed recurrent formulation. Retaining the better suited recurrent formulation of attention, in Experiment 2, we evaluate the improved encoder architecture along with a more advanced decoding method (beam-search).

\subsection{Evaluation metrics}
\label{sec:transf_similarity}

\paragraph{Bias in model evaluation using BLEU score}
While BLEU is a standard metric, it only captures surface correspondence between the references and generated text. In the dataset, a sequence of motion is mapped to multiple references. If we take the case of two samples which have the same sequence of motion approximately and different sets of references. Considering as example the  prediction \textit{a person wipes a table.}, and it references \textit{a human wipes on a table.},\textit{ a person stirs with their left hand.}. The prediction is counted correct only if the words are the same. A description with the same meaning and a different formulation will have low BLEU, which doesn't correlate with human judgment. Consequently, the BLEU score doesn’t compare semantic correspondence, to overcome this limitation we propose a semantic similarity metric, invariant under paraphrase, to evaluate the semantic adequacy of generated text compared to references. 

\paragraph{Sentence semantic similarity score}
We can devise a simple transformer-based metric, similar to a triplet-loss approach: by exploiting a sentence transformer \cite{reimers-2019-sentence-bert} model trained for paraphrase detection using triplet loss, we can compute a normalized cosine similarity score between each reference and the generated sentence.  Since movement descriptions are composed of very common general-domain vocabulary, a generalist sentence embedding model trained to assign high scores to vector embeddings of paraphrased sentences, can be expected to perform very well at evaluating the semantic equivalence of the generated sentence with regard to the references. 

\begin{equation}
        score(RC, P) = \frac{1}{\mid RC \mid}\sum_{i\in \mid RC \mid } sim(RC_i, P_i)
\end{equation}

\begin{equation}
    sim(RC_i, P_i) = \frac{1}{\mid P_i \mid}\sum_{p\in P_i} \max_{r \in RC_i} sim_{cos}(emb(p), emb(r))    
\end{equation}

Where RC is the reference corpus and where each \(r_i \in RC\) is a set containing multiple references, and where P is the predictions and where \(p_i \in P\) is a set containing one or more predictions. \(emb\) represents the sentence-embedding model and returns an embedding vector for a sentence or phrase and \(sim_{cos}\) is the normalized scalar product between the vectors. Since references can be quite different from one another, although both are correct, we only keep the highest score for a prediction with relation to all corresponding references. Here, the \(sim_{cos}\) is used because the sentence transformer paraphrase model used is trained using the cosine similarity as a regression loss. Other sentence-transformers models may require the use of different metrics. 

\begin{table}[b]
\centering

\begin{tabular}{ccc}
\toprule
\textbf{System} & \multicolumn{2}{c}{\textbf{Soft.att}}\\ 
 & \textbf{Similarity}  & \textbf{BLEU score} \\ 
\midrule
GRU-Angles& 79.50 & 28.5 \\ 
BiGRU-Angles& 78.90 & 26.3 \\ 
GRU-Cartesian & \textbf{79.56} & \textbf{31.8} \\ 
BiGRU-Cartesian& 78.84 & 28.6\\
\bottomrule
\end{tabular}
\caption{GRU-based system : Results of soft attention }
\label{tab:sent_eval_bleu_gru_gru}
\end{table}

\subsection{Experiment 1: Attentions effect using (Bi)GRU-as encoder}
\label{sec:gru_results}

\subsubsection*{Soft attention} 

We evaluate and compare the angle-based text generation model with the Cartesian coordinate generation models, both bidirectional and unidirectional. The angle joints in this split have allowed to attain a BLEU-4 of $26.3\%$, while the use of cartesian coordinates has allowed to increase this performance  up to $28.6\%$ on the test set ($+2.5\%$). Regarding the semantic evaluation presented in Table \ref{tab:sent_eval_bleu_gru_gru}, the difference is small, between predictions based on joint angles $78.90\%$ versus cartesian coordinates $78.84\%$. We can confirm this  in our error analyses. We note also that in most cases the predictions have a good semantic score ($>78\%$) while the BLEU score is under $32\%$, which doesn't represent a fair and realistic evaluation of generation quality or performance (see section \ref{sec:transf_similarity}).

\begin{table}[t]
\centering
        \begin{tabular}{cccccc}
        \toprule
        \textbf{System } & \textbf{Mask} & \multicolumn{2}{c}{\textbf{Similarity}}  & \multicolumn{2}{c}{\textbf{BLEU score}}\\
         & \textbf{(D=5)} & \textbf{Loc.rec.att} & \textbf{Loc.att}  & \textbf{Loc.rec.att} & \textbf{Loc.att}\\ 
        \midrule
        GRU-Cartesian & True & 78.84 &78.44 & 30.2 & \textbf{30.0} \\ 
        GRU-Cartesian & False & \textbf{80.54} & 78.34 & 29.7 & 29.3 \\ 
        BiGRU-Cartesian & True & 78.64 &79.53 & \textbf{31.1} & 29.3 \\ 
        BiGRU-Cartesian & False & 79.31 & \textbf{79.58} & 26.3 & 27.6 \\ 
        \bottomrule
        \end{tabular}
\caption{GRU-based system : Results of local attention and recurrent attention.}
\label{tab:bleu_sim_rec_gru_gru}
\end{table}

\subsubsection*{Local and recurrent attention}
\quad Table \ref{tab:bleu_sim_rec_gru_gru} presents the results of applying local attention and the extended recurrent $p_t$ computation. The highest BLEU score of $31.1\%$ is obtained for BiGRU + masked local recurrent attention. However, as demonstrated in section \ref{sec:limit_loc_soft}, this model does not perform any proper semantic segmentation contrarily to the case of a unidirectional GRU, obtaining a lower BLEU but a higher similarity score $80.54\%$ and showing a medium qualitative synchronization between motion and language. For local attention, the GRU seem to be the better model in terms of BLEU score $30\%$.

\subsection{Experiment 2: Efficacy of MLP-GRU}
\quad In this section we discuss the results resulting from using the simple MLP as the encoder network (cf.Figure \ref{fig:all_arch} design \textbf{3}) and applying local recurrent attention. Table \ref{tab:mlp_gru} summarizes the BLEU and similarity scores for all $MLP-GRU$ experiments. In all reported scores from the experiments, the hidden size is fixed at $d_i=64$. The results for a higher hidden size $d_{i} \in \{128,256\}$ were slightly lower and sometimes give approximately the same performances as $d_i=64$, but start to overfit rapidly. There is more marked qualitative difference through visual inspection of a sample, which was globally lower in the case of high dimension. Maximal BLEU and faster convergence was achieved with no mask and a high hidden size. With a mask and a lower hidden size, the training takes more time but the semantic semantic segmentation is better. 

\begin{table}[t]
\centering
\begin{tabular}{lcccc}
\toprule
\textbf{Design}       & \textbf{Mask} & \textbf{D} & \textbf{BLEU}  &  \textbf{Similarity}  \\ 
\midrule
MLP-GRU (L=4)         & True          & 5          & \textbf{32.1 } &        78.29             \\ 
MLP-GRU               & True          & 5          & \textbf{31.6 } &   \textbf{78.87}       \\
MLP-GRU               & False         & 5          & 28.8           &   78.29             \\ 
MLP-GRU               & True          & 9          &  29.1          & 78.52            \\ 
MLP-GRU               & False         & 9          & 28.4           &   76.38                  \\ 
\bottomrule
\end{tabular}
\caption{Results of MLP with $2$ layers, except for the case $(L=4)$ we use $4$ layers.}
\label{tab:mlp_gru}
\end{table}

\subsubsection*{Effect of the Mask window}

 When the mask is enabled, a truncated Gaussian window is applied in each decoding step $t$. When the mask is disabled, no truncation is performed, and  the alignment is bad in several experiments (see section \ref{sec:visanalysis}), along with a lower BLEU score ($<29\%$) as reported in the Table \ref{tab:mlp_gru}. The alignment wasn't possible without applying the mask, which acts as regularizer during training and reduces overfitting. The strict limitation of visible encoder outputs along with frame-level feature extraction pushes the network to forcibly learn a precise position only through a semantic motion segmentation based on the text. Another important factor is the $D$ value: for $D=9$ applying the mask doesn't affect the alignment, and in general when $D$ is high, making has almost no effect.
 
 The results of designs \textit{(1,2)} (Figure \ref{fig:all_arch}) presented in section \ref{sec:gru_results} rely more on global information as a simple way to give a correct prediction instead of learning motion segmentation (especially for short motion action). We explored also a \textit{causal} window given a mask range $\llbracket p_t-D,p_t \llbracket$ with $D=5$, but this led to a deteriorated BLEU score.

\begin{table*}[ht]
\centering
\resizebox{\textwidth}{!}{\begin{tabular}{c|l|l}

\textbf{Action} & \textbf{References} & \textbf{Predictions (beam size=3)} \\
\midrule
Kneeling & Human ducks down and takes her hands over her head & a person lies \textit{on all fours} and stands up \\
 & & a person lies \textit{on the ground} and stands up \\
 & & a person lies \textit{on the floor} and stands up \\ \hline
Stomping & Human is lifting his leg & a person is stomping with the right foot. \\
 & & a person is stomping with the left foot. \\
 & & a person is stomping with his right foot. \\ \hline
Squatting & A person stretches their arms out and performs \textit{a single squat} & a person performs a squat \\
 & A person does some squad exercises & a person performs a squat squat \\
 & & a person performs a single squat \\ \hline
Walking & Slow walking motion & a person walks 4 steps forward \\
 & A person walks forward slowly & a person walks \textit{slowly} 4 steps forward \\
 & Human slowly goes forward & a person walks forward \\ \hline
Wiping & A human wipes on a table & a person wipes something in front of it with the left hand \\
 & A person cleans a surface making circular motions with their left hand & a person wipes something in front of it \\
 & & a person wipes something in front of him \\ \hline
Waving & A person raises his right arm in front of his face then starts waving & a person waves with his left hand \\
 & Human performs a waving motion with right hand & a person waves with his right hand \\
 & A person waves with its right hand & a person waves with the left hand \\

\end{tabular}}
\caption{Predictions of model MLP-GRU (D=5) with a beam size of 3.}
\label{tab:beam_pred}
\end{table*}

\begin{table}[b]
\centering

 \begin{tabular}{ccccccc}
 \toprule
\multirow{2}{*}{\textbf{Model}} & \multirow{2}{*}{\textbf{D {[}Mask{]}}} & \multicolumn{5}{c}{\bf BLEU   score (beam size)}                                                                                                   \\ 
                                &                                        & \multicolumn{1}{c}{1}             & \multicolumn{1}{c}{2}    & \multicolumn{1}{c}{3}             & \multicolumn{1}{c}{4}             & 6    \\ \midrule
\multirow{4}{*}{MLP}            & 5 {[}True{]}                           & \multicolumn{1}{c}{\textbf{31.6}} & \multicolumn{1}{c}{30.6} & \multicolumn{1}{c}{30.5}          & \multicolumn{1}{c}{30.7}          & 30.6 \\ 
                                & 5 {[}False{]}                          & \multicolumn{1}{c}{28.8}          & \multicolumn{1}{c}{29.7} & \multicolumn{1}{c}{\textbf{30.7}} & \multicolumn{1}{c}{30.4}          & 30.7 \\ 
                                & 9 {[}True{]}                           & \multicolumn{1}{c}{29.1}          & \multicolumn{1}{c}{30.4} & \multicolumn{1}{c}{\textbf{30.4}} & \multicolumn{1}{c}{30.1}          & 30.1 \\ 
                                & 9 {[}False{]}                          & \multicolumn{1}{c}{28.4}          & \multicolumn{1}{c}{\textbf{29.2}} & \multicolumn{1}{c}{28.8}          & \multicolumn{1}{c}{28.9} & 28.6 \\ \midrule
Deep-MLP                        & 5 {[}True{]}                           & \multicolumn{1}{c}{\textbf{32.1}} & \multicolumn{1}{c}{29.4} & \multicolumn{1}{c}{29.1}          & \multicolumn{1}{c}{29.1}          & 29.8 \\ \bottomrule
\end{tabular}
        \caption{Impact of beam searching on MLP based models}
\label{tab:bleu_impact_beam_search}
\end{table}

\subsubsection*{Impact of Beam-search} 
 In this followup experiment, we study the impact of beam search on the BLEU@4 score and average sentence length for the MLP-GRU.
 In Table \ref{tab:beam_pred} we show predictions from different actions: we can see that the model has learned a coarse classification of different meanings \textit{(on all fours, on the ground, on the floor}). For the \textit{Stomping} and \textit{Waving} actions we see that the first and second prediction are focusing on which body part is executing the motion (left or right/ leg or hand) but the first prediction is correct. For the \textit{Walking} action, we see that the information about the speed of motion is only present in the second prediction. We confirm that applying the beam search for different beam size does not give an amelioration of the BLEU score for the models based on MLP. More results are reported in the Table \ref{tab:bleu_impact_beam_search}. The best BLEU score is already obtained without sampling, thus no need to perform a beam searching in the prediction process, which is expensive in time computation.

\subsection{Global comparison}
In summary, across the experiments, we observe that models based on recurrent encoders more directly rely on global information. The training for such architectures is easier and converges more quickly to higher BLEU scores. However, the distribution of attention weights along the frames axis doesn't match the intuitive semantic segmentation produced by human analysis. Introducing recurrent local attention and using a simple MLP as an encoder pushes the network to improve the BLEU score only by learning a correct synchronized semantic segmentation. The convergence for such designs takes longer as the decoder is limited to the visible encoder outputs (finding the correct position $p_t$ takes longer).

\subsection{Comparison with SOTA}
The comparison was challenging because different authors use different seeds to split the dataset (although proportions are the same), without reporting precisely which random seed is used. In addition to reporting the performance of SOTA systems, we also evaluate our models with three different seeds to verify that the variations of performance due to different random seeds is minimal (less than 0.01 difference), which allows some comparability with SOTA systems. The results are reported in the Table \ref{tab:mlp_vs_SOTA_beam_analyze}. Across all random splits, our BLEU score is higher than reported systems by a margin significantly larger than seed-related variations. 

 \begin{table}[ht]
\centering
\begin{tabular}{cccccc}
\toprule
\multirow{2}{*}{Random   state} & \multirow{2}{*}{Model} & \multicolumn{4}{c}{BLEU  score}                                                                     \\ 
                                &                        & \multicolumn{1}{c}{1st}   & \multicolumn{1}{c}{2nd}   & \multicolumn{1}{c}{3rd}   & Avg            \\ \midrule
\multirow{2}{*}{0}              & MLP                    & \multicolumn{1}{c}{0.286} & \multicolumn{1}{c}{0.272} & \multicolumn{1}{c}{0.267} & 0.275          \\  
                                & \textbf{DeepMLP}       & \multicolumn{1}{c}{0.286} & \multicolumn{1}{c}{0.298} & \multicolumn{1}{c}{0.262} & \textbf{0.282} \\ \midrule
\multirow{2}{*}{35}             & MLP                    & \multicolumn{1}{c}{0.291} & \multicolumn{1}{c}{0.274} & \multicolumn{1}{c}{0.267} & 0.277          \\ 
                                & \textbf{DeepMLP}       & \multicolumn{1}{c}{0.297} & \multicolumn{1}{c}{0.268} & \multicolumn{1}{c}{0.270} & \textbf{0.278} \\ \midrule
\multirow{2}{*}{42}             & \textbf{MLP}           & \multicolumn{1}{c}{0.304} & \multicolumn{1}{c}{0.286} & \multicolumn{1}{c}{0.265} & \textbf{0.285} \\
                                & DeepMLP                & \multicolumn{1}{c}{0.256} & \multicolumn{1}{c}{0.265} & \multicolumn{1}{c}{0.255} & 0.259          \\ \midrule
\_ & GASSL(LSTM)\cite{Goutsu2021} &   \multicolumn{1}{c} {0.288} &  \multicolumn{1}{c} {0.263} & \multicolumn{1}{c} {0.257} & \multicolumn{1}{c}{\textit{0.269}} \\ 
\_ & GASSL(GRU)\cite{Goutsu2021} &   \multicolumn{1}{c} {0.262} &  \multicolumn{1}{c} {0.243} & \multicolumn{1}{c} {0.231} & \multicolumn{1}{c}{\textit{0.245}} \\ 
\_ & BiLSTM \cite{Plappert2017} &   \multicolumn{1}{c} {0.259} &  \multicolumn{1}{c} {0.237} & \multicolumn{1}{c} {0.249} & \multicolumn{1}{c}{\textit{0.248}}       \\ \bottomrule
\end{tabular}
\caption{[Original KIT-ML]Comparison with other models on different level of beam searching.}
\label{tab:mlp_vs_SOTA_beam_analyze}
\end{table}

\subsection{Experiment 3: Semantic Segmentation Evaluation}

For qualitative evaluation of primitive segmentation, we label some representative samples in the dataset for every action (\textit{Kicking}, \textit{Walking}, \textit{Stomping}…). For evaluation, we need an adapted alignment score, this process required to introduce precise definition of a motion segment. As detailed in Lin et al. \cite{Lin2016}, the definition is specific to the task. In our work, we want to evaluate both alignment and synchronization between description words and motion primitives. For this purpose we propose three adapted method to calculate a segmentation score \textit{i) based on intersection over union, ii) based on intersection over prediction and iii) based only on alignment position}. The main goal of this study is to quantitatively verify the accuracy of synchronization between correctly generated descriptions and motions: we chose to not evaluate the synchronicity for a completely wrong descriptions, which can never be synchronised. The evaluation of synchronicity is a process of verification that comes after a correct semantic prediction.

\subsection{Results on HumanML3D and augmented KIT-ML}

To further investigate the generalization capability of our method, we conduct additional experiments on the updated version of KIT-ML, which was augmented by mirroring provided in \citep{Guo_2022_CVPR}. Moreover, we adapt the proposed architecture to deal with much larger dataset such as HumanML3D \citep{Guo_2022_CVPR}.

\paragraph{Training configuration.}
We use the same split on which we conducted all previous alignment experiments. Later, after our work, an augmented version of the KIT dataset was released by \citep{Guo_2022_CVPR}. Consequently, we retrained our models and compared them with the state-of-the-art model released by \citep{chuan2022tm2t}.

\textit{KIT-ML \citep{Guo_2022_CVPR}.} The model hyperparameters remain the same for this augmented version.

\textit{HumanML3D \citep{Guo_2022_CVPR}.} For this larger dataset, the word embedding dimension is set to $128$ and the hidden size to $d_i = 256$, while the hidden dimension of the \verb|MLP| with 2 layers have respectively an output dimensions of $512$ and $256$ (represent $d_{enc}$) and $D=10$.

\paragraph{Quantitative comparison vs SOTAs.} In the Table \ref{tab:transf_vs_gru}, we report BLEU scores for different splits and dataset versions, comparing the Transformer and GRU-based generation. Our model based on the proposed local recurrent attention achieve better performance than the transformer based model \verb|TM2T| and more importantly we preserve the semantic segmentation capability.

\begin{table}[t]
    \centering
     \resizebox{\linewidth}{!}{
        \begin{tabular}{lccccccc}
                    \toprule
                    \textbf{Dataset} &   \textbf{Model}  & \textbf{BLEU@1} & \textbf{BLEU@4} & \textbf{CIDEr} & \textbf{ROUGE-L} & \textbf{BERTScore}  & \textbf{Segmentation} \\ \midrule 
        
                    \multirow{2}{*}{KIT-ML}   & TM2T \citep{chuan2022tm2t} &  46.7      & 18.4 & 44.2 & 79.5     & 23.0  & \ding{55}\\ 
                                               & Ours MLP+GRU & \textbf{56.8} & \textbf{25.4} & \textbf{125.7} & \textbf{58.8} & \textbf{42.1}   & \ding{51} \\ \midrule
                                                 
                    \multirow{2}{*}{HumanML3D}   & TM2T \citep{chuan2022tm2t}   &  61.7      & 22.3 & 49.2 & \textbf{72.5}   & \textbf{37.8} & \ding{55} \\ 
                                                &  Ours|MLP+GRU| & \textbf{67.0} & \textbf{23.4} & \textbf{53.7} & 53.8 & 37.2 & \ding{51} \\
                    
                   \bottomrule 
        \end{tabular}}
        \caption{Comparison with SOTA results on the augmented version of KIT-MLD and HumanML3D.}
        \label{tab:transf_vs_gru}            
\end{table}

\subsubsection{Segmentation definitions} 
First, we will lay a few definitions before delving into the evaluation scores. 

\textbf{Word motion segment} The motion segment corresponding to the motion phase associated with the \textit{word} $w_i$ from a generated description will be denoted $S_i$  (as defined by Equation \ref{eq:seg_def}).

\textbf{Primitive motion} A basic motion characterized by a unique combination of \textit{action}, \textit{direction}, and \textit{speed}. A global change in the motion direction is considered as a \textit{primitive transition motion}.

\textbf{Language segment}
A linguistic description segment is the phrase associated with the word describing one specific action. As actions are almost always described by verbs, the linguistic description segments associated are almost always verbal phrases.

The two process given the motion primitive and the corresponding language segment will be referred respectively as motion and language segmentation. The details of these segmentations will be formulated in the section below.

In all following Equations, we note $S_{seg}^e$ the segmentation score of a sample $e$, and $\mid E \mid$ the cardinality of the set $E$. We note by $\mathbb{1}_{A}$ the indicator function ($\mathbb{1}_{A}=1$ if condition $A$ is true and $0$ otherwise).

\subsubsection{Methods for motion and language segmentation}

\textbf{Language segmentation} To perform language segmentation, we first retrieve the words corresponding to the actions from the annotations and calculate their indexes $k_m$ in the prediction. Let $w_m$ be the \textit{action word annotation} of a motion segment, we search for the index $k_m$ of the word $w_m$ in the predicted words then for the next annotation $w_{m+1}$ word and so forth until the final primitive segment is reached, we will write the index of the end token as $k_e$ . As a result of this process, formally, a language segment $L_{m}$ is given by $\bm{L_{m}=\{w_r, r \in \llbracket k_m,k_{m+1}-1 \rrbracket \}}$.

\textbf{Motion segmentation} We define $P_m$  as the segment of a primitive motion. The calculation of $P_m$ is given by the concatenation of words motion segments according to index in the set, $\llbracket k_m,k_{m+1}-1\rrbracket$ as given by the Equation \ref{eq:seg_primitive_motion}. We recall that the interval of the word motion segment $S_i$ was given in the Equation \ref{eq:seg_def}.

\begin{equation}
    \centering
     P_{m} = \bigcup_{i=k_m}^{k_{m+1}-1}{S_{i}}
   \label{eq:seg_primitive_motion}
\end{equation}

For a qualitative evaluation of motion and language synchronization, we propose to calculate the segmentation score using three different methods.

\textit{i) Intersection over Union (IoU)} Measure the proportion of intersection between the primitive segment and a reference segment. Based on the $IoU$ we calculate the segmentation score by the Equation \ref{eq:seg_score}.

\begin{equation}
    S_{seg}^e = \frac{1}{n_s} \sum_{k=0}^{n_s-1}{\mathbb{1}_{(\iouk \geq \theta)}}  , \   \iouk_{k \in [0,n_s[} = \frac{\mid P_{k} \bigcap G_{k} \mid}{\mid P_{k} \bigcup G_{k}\mid}  
\label{eq:seg_score} 
\end{equation} 

we note by $n_s$ the number of motion segment, $m \in \llbracket 0,n_s\llbracket$ is the motion primitive index, $\theta$ is the selected threshold, and $k_m$ computed by the language segmentation.
The reference motion segment for an \textit{action word} $w_k$ is the ground truth segment $G_k$, represent also the $k^{th}$ primitive for a given motion sample. 
The measure $IoU$ is sensitive to the absolute quality of the ground truth, which isn't always desirable due to variability, which led us to also propose the following measures.

\textit{ii) Intersection over Prediction (IoP)} We introduce the $IoP$ as a measure for the inclusion proportion of $P_k$ in $G_k$, so that we have $\bm{IoP_k=1 \Leftrightarrow P_k \subset G_k} $, meaning that the segmentation is counted as completely correct when the prediction segment interval is in the range of the ground truth interval. This measure correlates well with the perceived visual quality of transcription generation and reduce the impact of annotation subjectivity (start/end primitive annotation uncertainty). In this case, $S_{seg}^e$ is calculated as in Equation \eqref{eq:seg_score_1}.

\begin{equation}
\centering
S_{seg}^e = \frac{1}{n_s} \sum_{k=0}^{n_s-1}{\mathbb{1}_{(\iopk \geq \theta)}}, \ \iopk_{k \in [0,n_s[} = \frac{\mid P_{k} \bigcap G_{k}\mid}{\mid P_{k} \mid}
\label{eq:seg_score_1}
\end{equation}

\textit{iii) Element of} Evaluate if the alignment position $p_i$ is an element of the annotated segment interval. The score $S_{seg}^e$ is directly defined by \eqref{eq:seg_score_2}. This is the least strict method counting any amount of overlap as correspondence.

\begin{equation} \label{eq:seg_score_2}
    S_{seg}^e =  \frac{1}{n_s} \sum_{m=0}^{n_s-1} \sum_{i=k_m}^{k_{m+1}-1}{\mathbb{1}_{(p_i \in G_i )}}
\end{equation}

\begin{figure}[ht]
    \centering
    \includegraphics[width=.9\columnwidth]{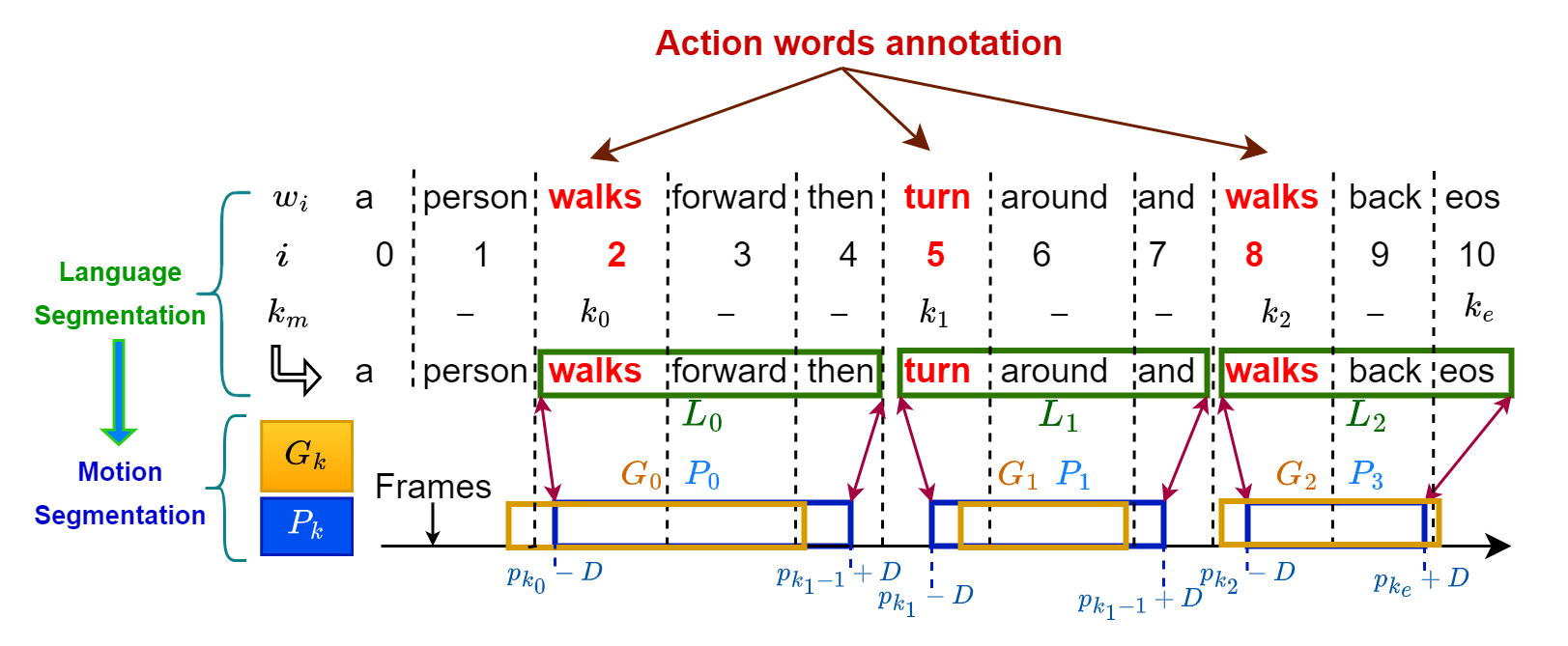}
    \caption{ The language and segmentation process applied on an example of predicted words $w_i$ with given action words  and frame annotations $G_k$. The inferred motion segment is $P_k$ and the language segment is $L_k$. The index $k_e$ always refers to the end token.}
    \label{fig:graph_segm}
\end{figure}

\paragraph{Segmentation process}
The two processes of language and motion segmentation are illustrated in Fig.\ref{fig:graph_segm}. Given a motion $x$ and the predicted description $y_x$ as \textit{“a person walks forward then turn around and walks backs} \verb-<eos>-. Let $\{S_i, i\in \llbracket 0,10\rrbracket \}$ be the set of motion segments associated to the eleven words forming the description $y_x$. Using the definitions above, we have three mappings :
\newline
$P_0 \rightarrow \ S_{k_0},...,S_{k_{0-1}} \rightarrow$ \textit{walks forwared then}~; 
 $P_1 \rightarrow$ \textit{turns around and}~; $P_2 \rightarrow$ \textit{walks back} \verb-<eos>-.
\newline
The indices $k_m$ are found by searching the ground truth word describing the action in the predicted sentence, and then using the corresponding segments for the group of words describing that part of the (Figure \ref{fig:graph_segm}). For this specific example, ($k_0=2,k_1=5,k_2=8)$. Consequently, the motion $x$ is a sequential composition of three primitive motions $P_0,P_1$ and $P_2$. Ideally $G_i=P_i$ but the subjectivity of human annotation introduces an inevitable variability in the start/end timestamps. Note that we always include the end token \verb-<eos>- in the final language segment, because it's relate to the end of motion.

\begin{figure*}[t]
    \centering
    \includegraphics[width=0.95\textwidth]{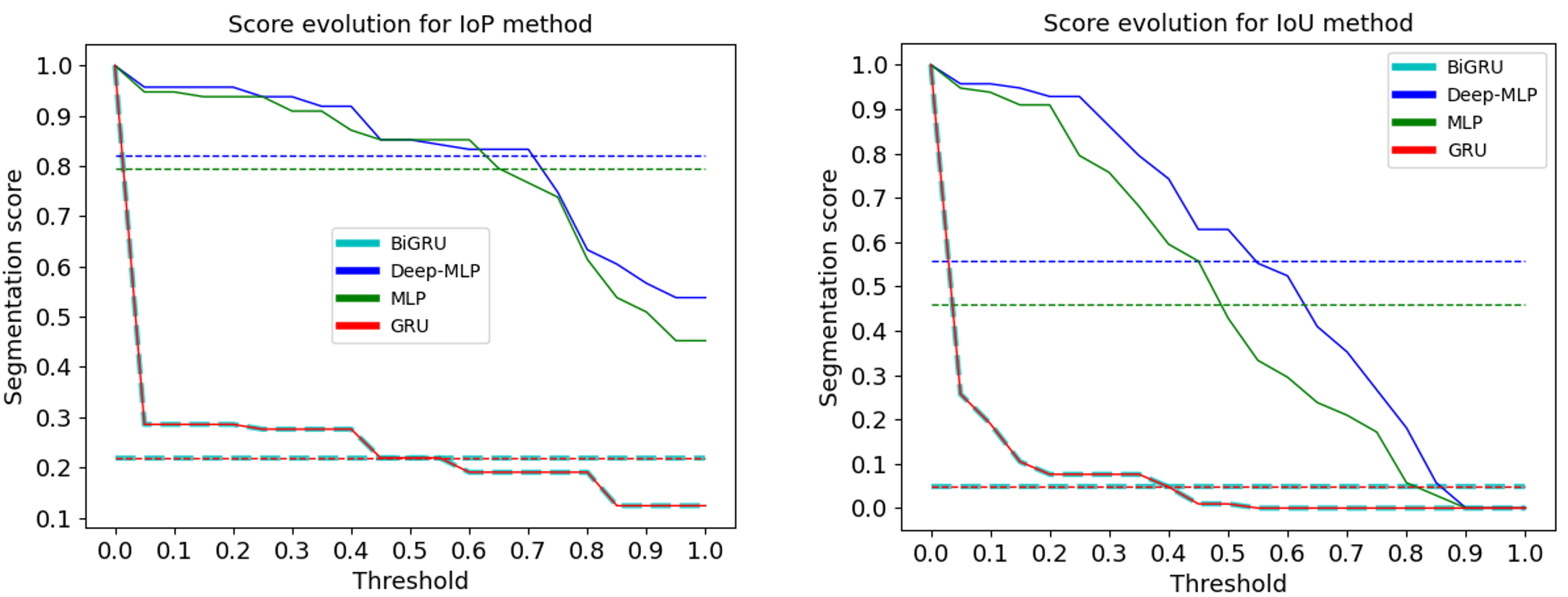}
    \caption{Comparison of IoP and IoU scoring methods. The legend specifies the encoder types.  The dashed lines represent the continuous average value without applying the threshold.}
    \label{fig:segScore_vs_th}
\end{figure*}

\subsubsection{Quantitative evaluation}
 
We have manually annotated 68 samples from KIT-ML with \textit{motion segments} and their \textit{action words}. For a compatible comparison between models, we select the samples with a semantically correct prediction in common across the models. This selection results in  \textbf{$N=35$ }common correct predicted representative samples for the calculation of scores. 

Each score is the mean value calculated on all samples $\frac{1}{N}\sum_{e=0}^{N-1}S_{seg}^e$. The scores as defined previously are computed with a threshold, anything below is zero, anything above is one. We can also compute a continuous variant, where we sum the actual scores without thresholding: $S_{seg}^e = \frac{1}{n_s} \sum_{k=0}^{n_s-1}{m_k}$. Fig.\ref{fig:segScore_vs_th} illustrates the thresholded (solid lines) score for thresholds from 0 to 1 as well as the continuous scores (dashed lines). \emph{Element of} is not included as it isn't compatible with such a visualisation. As expected the $IoU$ is too strict and the score is almost always zero even when the synchronisation is perceived as correct through visual inspection. We conclude that this measure is not adequate to give a fair assessment of how humans perceive synchronisation. We also observe that the continuous scores remove the parameter that is the threshold and give an informative assessment of synchronisation quality. 

In Table \ref{tab:seg_res}, we report all continuous scores, including \emph{element of} for a final comparison. We see that the segmentation performance of the GRU was low in comparison with the MLP encoder. The BiGRU has approximately the same performance as GRU. The two have a problem of a late detection of motion. The deeper MLP seems slightly better in segmentation than the shallow MLP regarding the continuous values $IoU$ and $IoP$. Which is confirmed by the respective curves (Figure \ref{fig:segScore_vs_th}). Noting also that the curves for GRU and BiGRU decrease exponentially w.r.t the threshold, confirming the low ability of recurrent encoders to perform semantic segmentation. Regarding the metric based on \textit{element of}, the MLP encoder is slightly better than Deep-MLP. This means that the MLP encoder localizes the time where the action happens better, while Deep-MLP localizes the start/end of the complete action better.

\begin{table}[ht]
    \centering
   \begin{tabular}{cccc}
   \toprule
        \textbf{System}       & \textbf{IoU}   &  \textbf{IoP}    & \textbf{Element of }  \\ \midrule
        \textbf{MLP-GRU}      & 45.97 &  79.47  & \textbf{89.05 }     \\ 
        \textbf{MLP-GRU (Deep)} & \textbf{55.92} &  \textbf{82.06}  & 88.09       \\ 
        \textbf{GRU-GRU  }    & 5.01 &  21.90  & 6.67 \\ 
        \textbf{BiGRU-GRU  }    & 5.01 &  21.90  & 6.67 \\ \bottomrule
    \end{tabular}
    \caption{Continuous segmentation scores for four encoder type [MLP, Deep-MLP, GRU, BiGRU]. All with Cartesian coordinates and loc.rec.att [D=5,Mask=True]}
    \label{tab:seg_res}
\end{table}

\section{Visual analysis} 

\subsection{Mapping words attention and motion}

\label{sec:visanalysis}
\begin{figure}
    \centering
    \includegraphics[width=0.8\columnwidth]{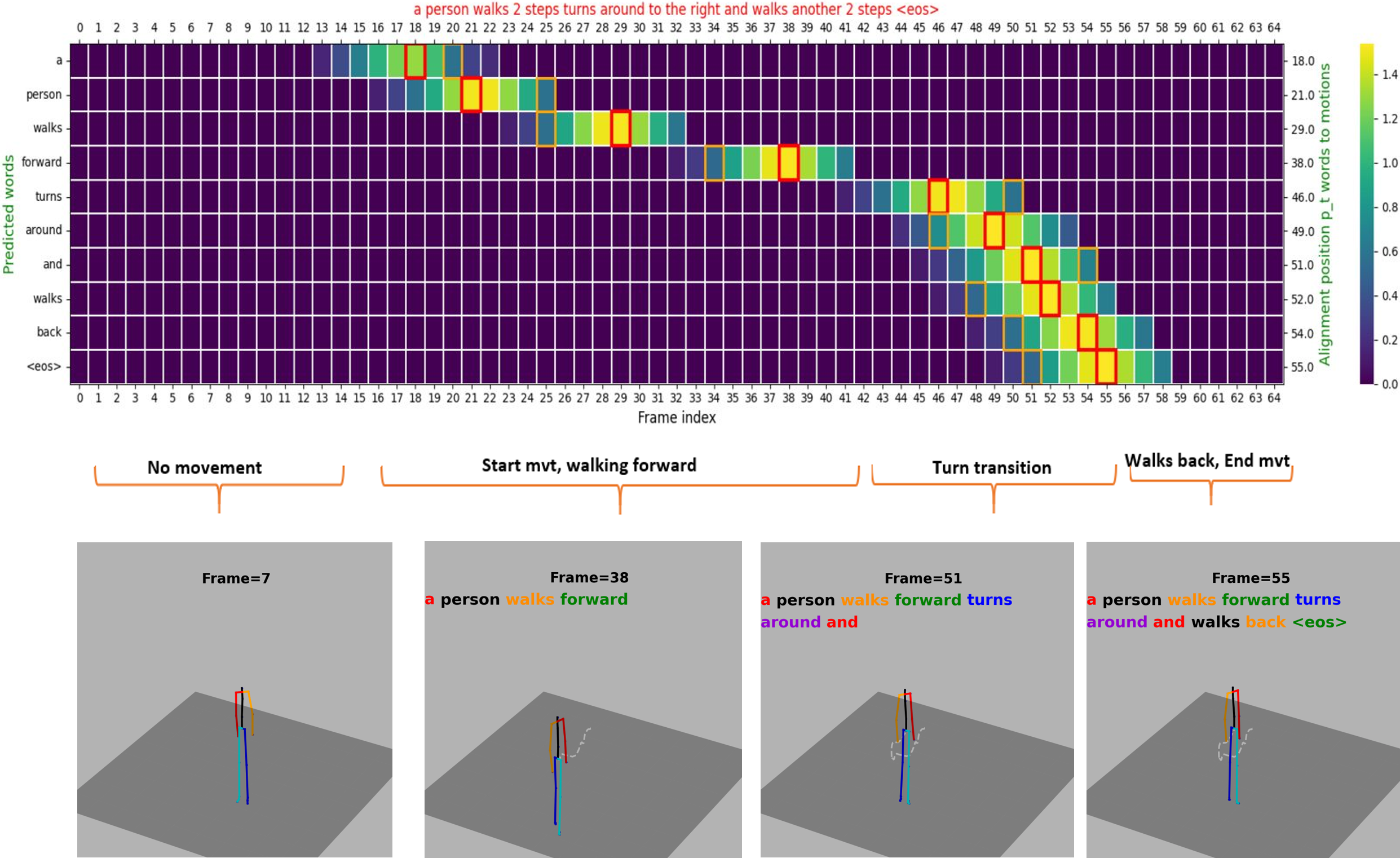}
    \caption{MLP-GRU [Local.rec.att Mask True $D=5$]: walk forward in the range $\llbracket 15,38 \rrbracket$, turn at frame $40$ , walk backward in $\llbracket 42,53\rrbracket$. Another animation illustrating synchronization between text and motion can be found in our code repository.}
    \label{fig:trunc_g_}
\end{figure}

In this part, we will discuss the perceived synchronization of generated words with motion through a qualitative error analysis. We illustrate this using the evolution of human skeleton motion and attention maps. The Fig.\ref{fig:trunc_g_} shows a correct segmentation in the presence of multiple primitive motions (U-turn). This motion is composed of three primitives "walks forward", "turns around" then "walks back", the primitive "turns around" represents the transition action.

\begin{figure}[h]
    \centering
     \includegraphics[width=\linewidth]{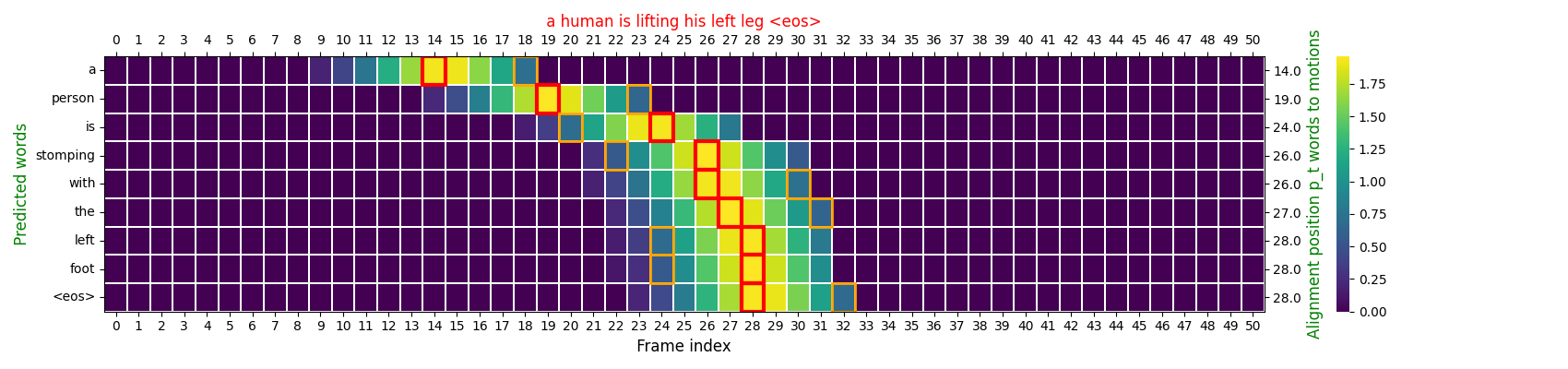}
    \caption{Truncated Gaussian MLP-GRU $D=5$, Stomping action at $\llbracket 22,30 \rrbracket$.}
    \label{fig:att_132_}
\end{figure}

 The Fig.\ref{fig:att_132_} shows a stomping action, where the word \textit{“stomping”} is associated to the frame interval $\llbracket 22,30 \rrbracket$, which is the exact moment of execution of the stomping action, followed by an accurate recognition of the body part executing the motion \textit{“left foot”}. Note that the reference and the generated descriptions are perfect paraphrases, they convey the same meaning but use different words. The BLEU score would be incorrectly low in this case. Also note that the attention weight start to be higher when the human skeleton starts moving around the frames in the range $\llbracket 13,16 \rrbracket$. 

\FloatBarrier
\subsection{Synchronization between motion and words}
\begin{figure}[h]
    \centering
    \includegraphics[width=0.7\columnwidth]{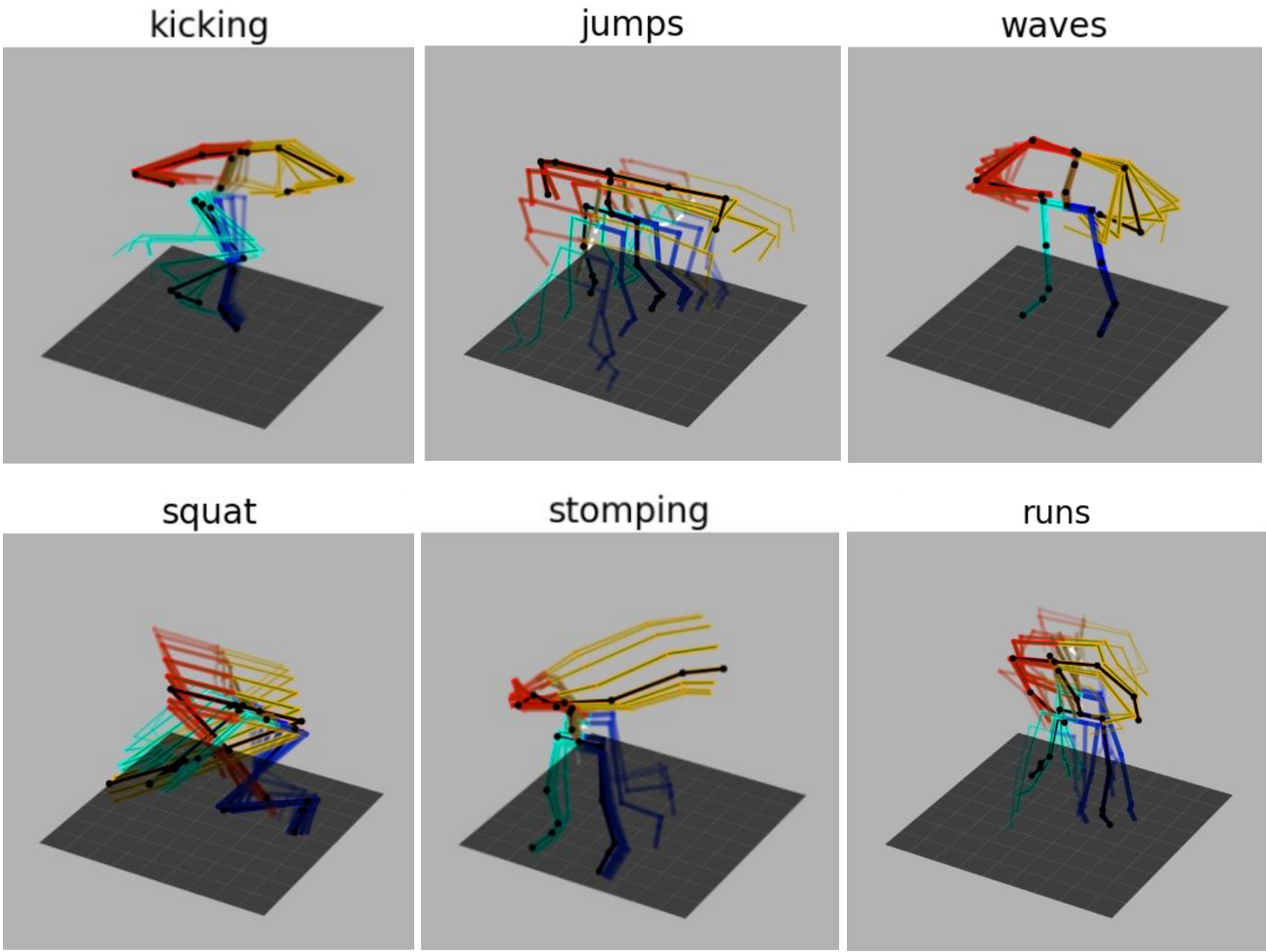}
    \caption{MLP-GRU [Local.rec.att Mask True $D=5$]: action words and their high attention motion frames. Given a word $w_i$, the transparency of each frame $j$ is set to value $V_{ij}$ as described in Equation \ref{eq:transparency_equation} .}
    \label{fig:multi_w2M}
\end{figure}

We propose a better visualization method for the synchronization by a parameterized skeleton transparency using attention weights. A sequence of motion frame $j \in \llbracket 0,T_x-1 \rrbracket$ is generated with this transparency values for each predicted word $w_i$, where the transparency $V_{ij}$ for a frame $j$ given a word $w_i$ is computed from the attention weights using Equation \ref{eq:transparency_equation}. The motion sequence is shown for action words, as illustrated in Fig.\ref{fig:multi_w2M}. 

\begin{equation}
    \centering
    V_{ij} = \frac{1}{\max_{j<T_x}\{G_{ij}\}} \times G_{ij} \hspace{10mm} G_{ij} = \frac{\exp{(F \times \alpha_{ij})}}{\sum_{j=0}^{T_x-1}{\exp{(F \times \alpha_{ij})}}}
    \label{eq:transparency_equation}
\end{equation}

 The factor $F$ is set to $100$ for better visualization and controls the number of frames present per image. Note that none of these operations change the position of alignment and maximum of attention, it only helps to have a better visualization of synchronization as a frozen sequence inside the interval $\llbracket p_t-D,p_t+D\llbracket$ (mask=True). 3D animation can be found in our repository as a point of comparison to see the relevance of these static visualizations. The visualization of sequence skeleton may be only relevant for words directly describing the motion. We have found that for the repeated words \textit{“a person”} it is usually aligned with the start of movement, providing an initial position to predict the coming action.
 
\subsection{Mapping between language and motion segment}

\begin{figure}[t]
    \subfloat[Single action.\label{fig:lm_to_pm1}]{%
      \includegraphics[width=.472\columnwidth]{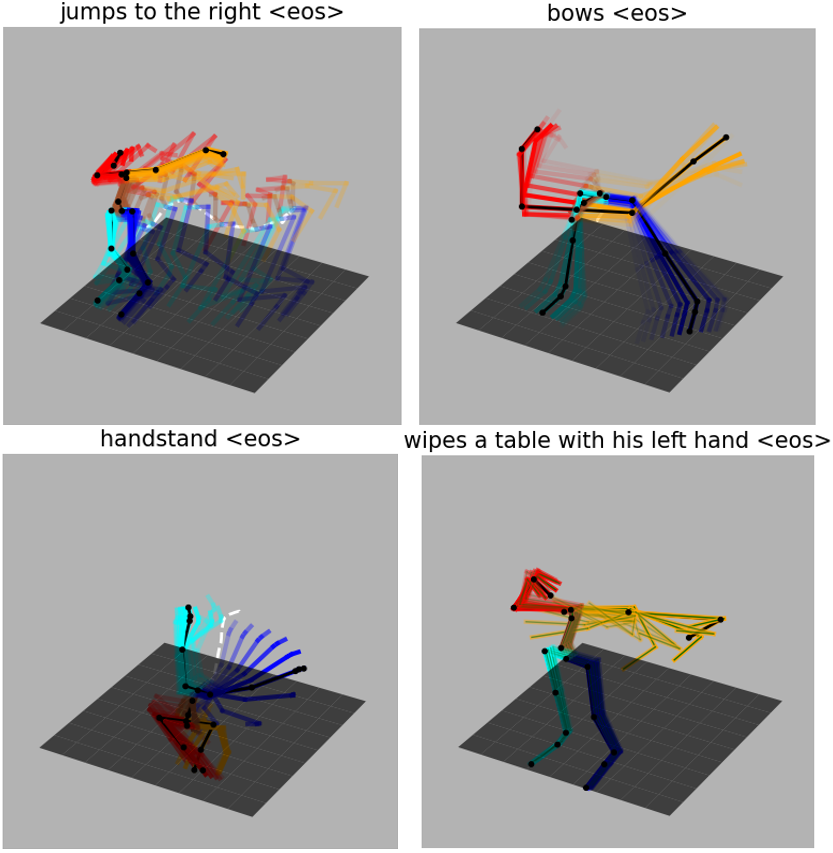}
    }
    \hfill
\subfloat[Multiple actions.\label{fig:lm_to_pm2}]{%
      \includegraphics[width=.45\columnwidth]{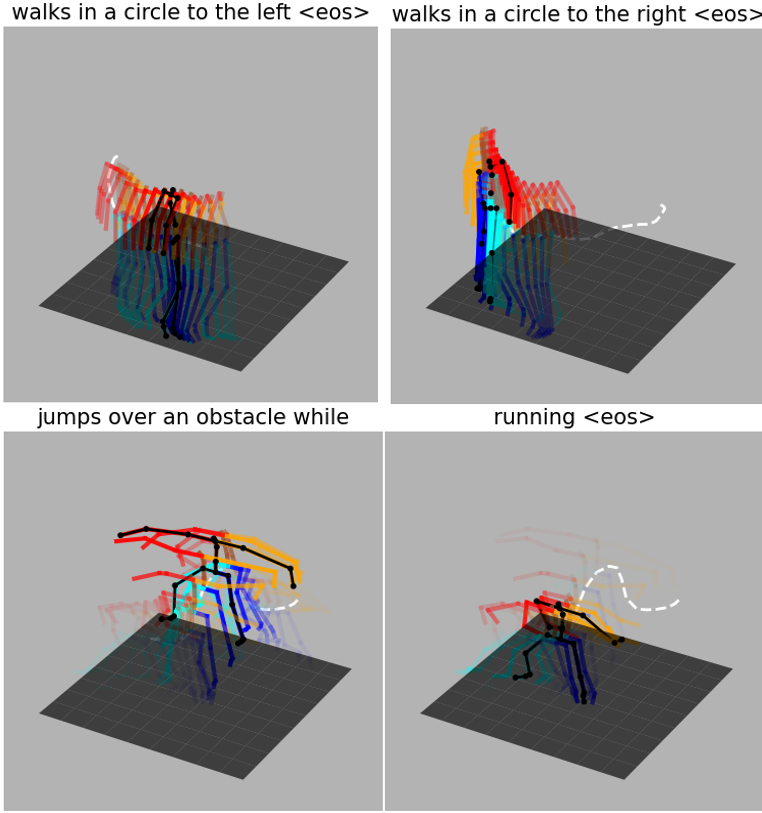}
    }
    \caption{Mapping $P_m$ \textrightarrow $L_m$: multiple actions vs. single action.}
    \label{fig:lm_to_pm}
  \end{figure}

As discussed before, it is more pertinent to associate a set of frames not only with a single action word but with the phrase describing the motion with higher granularity, which differs from mere action recognition. The Fig.\ref{fig:lm_to_pm} show the mapping of the human pose sequence for the frames in the set $P_m$ described by the language segment $L_m$. The transparency is calculated as before in Equation \ref{eq:transparency_equation}, but the coefficients $\alpha_{ij}$ are replaced by the coefficients $\gamma_{ij}$ defined in Equation \ref{eq:seg_att_coeff}, computed for each language segment $L_m$ ($m \in \llbracket 0,n_s-1 \rrbracket$).

\begin{equation}
 \gamma_{k_{m}j} = \frac{1}{k_{m+1}-k_{m}} \sum_{i=k_m}^{k_{m+1}-1}{\alpha_{ij}} \hspace{8mm} \forall m \in \llbracket 0,n_s-1 \rrbracket   
 \label{eq:seg_att_coeff}
\end{equation}

Specially when $m=n_s-1$ we should have $k_{m+1}-1 = k_{e}$. As described before, we always include the token \verb-<eos>- in the final language segment, so by definition $\bm{k_{n_{s}}=k_e+1}$.

\subsection{Limitations of the recurrent attention and dataset}

\quad \textit{Counting motion}
The MLD dataset doesn't contain sufficient examples with a variable number of repetitions to learn the invariants necessary for counting repetitions (number of steps, waving hands, \dots). Indeed, in the majority of cases, it includes only a few examples given fixed counts  \emph{(walking 4 steps)}. Instead, the architecture needs to see examples with a variable number of repetitions in the training data \emph{(e.g., walks 5,4,2 steps forward)} to allow for a better generalization. However, this is not the only requirement, a strict limitation of visibility of the input in the encoder will not allow the model to learn how to count repetitive motions, as a larger window width would be required. A partial solution can be to apply multiple Gaussian windows that could get counting information.

\textit{Dynamic motion measure} The information about the speed of the movements was seldom present in the dataset and consequently, the models are rarely able to generate descriptions regarding the speed of movements. 

\textit{Body part identification} We have seen in some samples a wrong detection of the body part executing the motion (e.g. left leg instead of right leg), which is the result of i) few descriptions rich enough to make the distinction in dataset ii) the limitations of the extracted features that do not take skeleton geometry into account.
Applying architectures allowing spacial feature extraction can be more pertinent to add this capability. 

\section{Conclusions and Perspectives}
We have proposed a supervised motion-to-language translation system that incorporates non-supervised alignment techniques based on attention weights that allow for synchronous or online text generation. Such a system, beyond the generation of motion descriptions in natural language, can be potentially adapted to other tasks where synchronous generation is important: subtitles of movies, sign language transcription, skeleton-based action segmentation, etc.
In general, multiple levers can be leveraged to drive the adaptation to adjacent tasks: adjusting the value of the $D$ parameter, tuning $\bm{\epsilon}$ to control the amount of overlap in the alignments (negative values are permissible). On the tasks of motion-to-language translation, there are additional improvements that may be investigated to further improve alignment performance. 
We believe that further developments in this area of motion-to-text translation placed in a larger scope, can bring about numerous potential applications, not just in social robotics or human-robot interaction. 
Motion segmentation approaches on their own have a myriad of potential applications \cite{Wang2018Applications}, but their intersection with description generation permits more focused applications requiring both qualitative description/assessment and time-synchronous delimitation:  ranging from sports dialectics and athletic performance optimization to functional rehabilitation medicine.

\section*{Data availability statement}
The KIT human motion to language dataset used throughout this work is available at \url{https://motion-annotation.humanoids.kit.edu/dataset/}. The HumanML3D dataset is available at \url{https://github.com/EricGuo5513/HumanML3D}. Any scripts performing post-processing of said data before it was used, as well as the portion we annotated with synchronization information, and the full implementation of our models, is available on our git repository \url{https://github.com/rd20karim/M2T-Segmentation/tree/main}. 

\section*{Acknowledgements} The following work is supported by the scholarship granted by the Occitanie Region of France (Grant number ALDOCT-001100 20007383). This version of the article has been accepted for publication, after peer review  but is not the Version of Record and does not reflect post-acceptance improvements, or any corrections. The Version of Record is available online at: \url{https://link.springer.com/article/10.1007/s00521-023-09227-z }.

\bibliographystyle{unsrtnat}
\bibliography{main}

\newpage

\section*{Appendix A}

In all conducted experiments the common observation is that most motion descriptions start with the words “a person”, for such words that depend on language model, the network learns to align this word with the start of the motion.

\FloatBarrier

\begin{figure}[ht]
    \centering
    \includegraphics[width=\columnwidth]{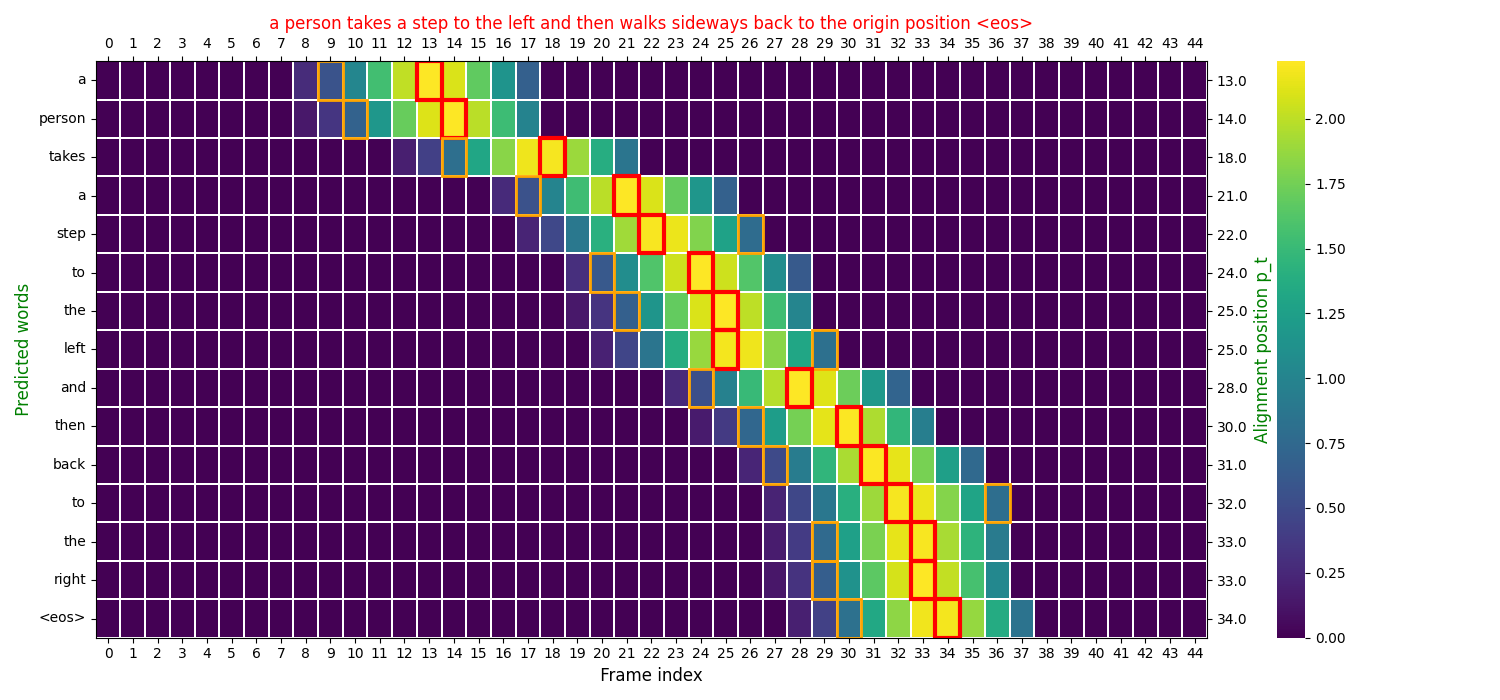}
    \caption{Truncated gaussian DeepMLP-GRU $D=5$ [move left, move right].(move to the left in the range $\llbracket8,23 \rrbracket$, move to the  right in the range $\llbracket24,35 \rrbracket$}
    \label{fig:left_and_right}
\end{figure}

Fig.\ref{fig:left_and_right} shows the case of a motion composed of two motion 'move to the left' then 'move to the right' where localization of attention weights was correctly distributed allowing a correct segmentation of motion.

\begin{figure}[ht]
    \centering
    \includegraphics[width=\columnwidth]{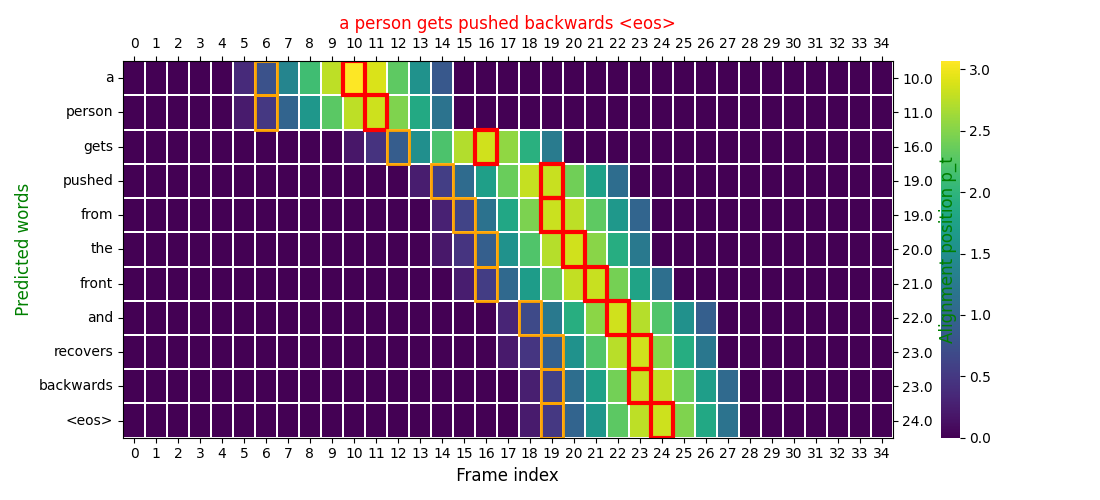}
    \caption{Truncated gaussian MLP-GRU $D=5$ [Push backward].(pushing action in the range $\llbracket15,26 \rrbracket$.}
    \label{fig:push_212}
\end{figure}

\FloatBarrier
Fig.\ref{fig:push_212} shows the case of push backward action, the attention positions were correctly distributed with relation to the range of the action.

\FloatBarrier

\begin{figure}[ht]
    \centering
    \includegraphics[width=\columnwidth]{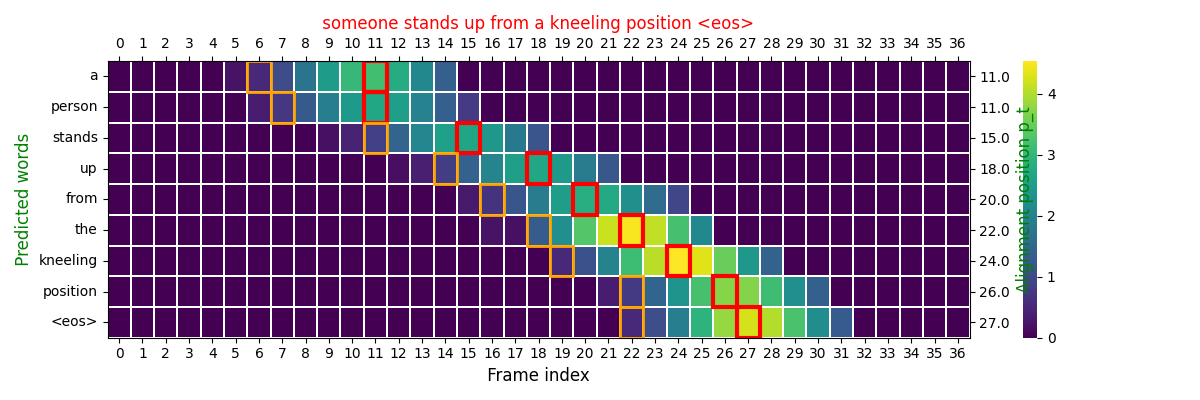}
    \caption{Truncated gaussian DeepMLP-GRU $D=5$ [stand up].(range $\llbracket11,30 \rrbracket$) }
    \label{fig:stand_up}
\end{figure}

Fig.\ref{fig:stand_up} represents the attention map for "walk down", when observing the motion, the action is performed in the range $\llbracket11,30 \rrbracket$ .

\begin{figure}[ht]
    \centering
    \includegraphics[scale=0.40]{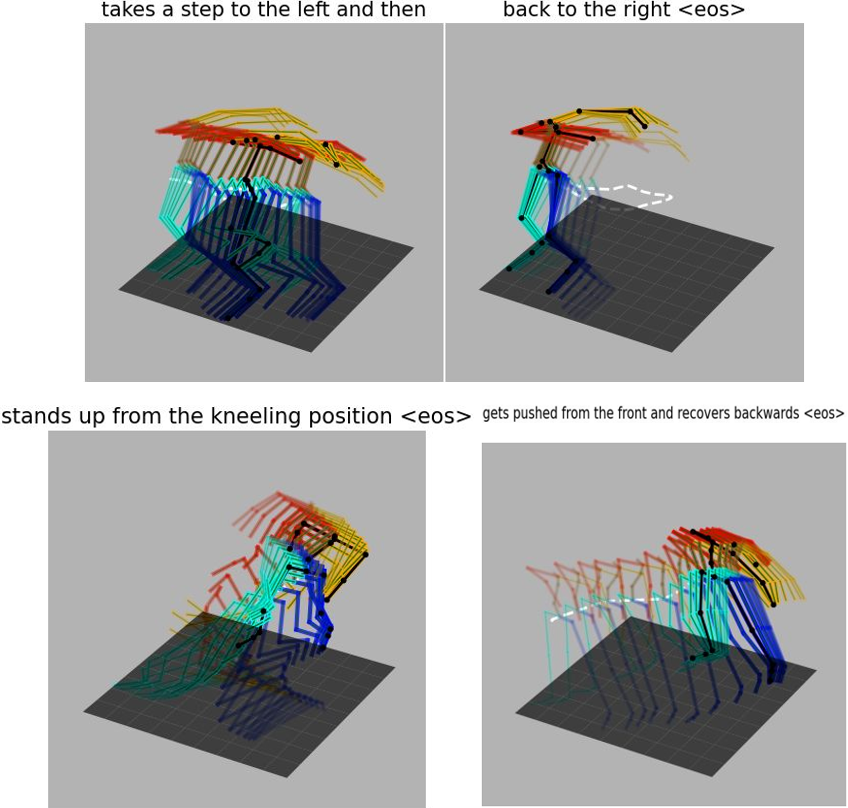}
    \caption{Truncated gaussian DeepMLP-GRU $D=5$ mapping $P_m$ \textrightarrow $L_m$ }
    \label{fig:frozen_deepMLP}
\end{figure}

\FloatBarrier
Fig.\ref{fig:frozen_deepMLP}, presents the association $P_m$ to $L_m$ for model using the deep-MLP encoder.

\end{document}